%% file: main.tex
\documentclass{article}
\usepackage{arxiv}
\usepackage{natbib}
\usepackage{graphicx} % Required for inserting images

\usepackage{amsmath,amssymb}
\usepackage{tikz}
\usetikzlibrary{calc}
\usepackage{xcolor}
\usepackage{comment}
\usepackage{amsthm}
\usepackage{wrapfig}
\usepackage{booktabs}
\usepackage{algorithm}
\usepackage{algorithmicx}
\usepackage[noend]{algpseudocode}
\usepackage{mathtools}
\usepackage{appendix}
\usepackage[T1]{fontenc}
\usetikzlibrary{positioning}
\usetikzlibrary{fit}
\usetikzlibrary{arrows}
\usetikzlibrary{decorations.pathreplacing}
\DeclareMathOperator{\E}{\mathbb{E}}
\input{math_commands.tex}

\usepackage[utf8]{inputenc} % Enable UTF-8 encoding

\usepackage{hyperref}
\usepackage{url}

\usepackage{titlesec}
\titlespacing\section{0pt}{0pt plus 0pt minus 2pt}{0pt plus 0pt minus 2pt}
\titlespacing\subsection{0pt}{0pt plus 0pt minus 2pt}{0pt plus 0pt minus 2pt}
\titlespacing\subsubsection{0pt}{0pt plus 0pt minus 2pt}{0pt plus 0pt minus 2pt}

\newtheorem{remark}{Remark}

\newtheorem{definition}{Definition}

\newtheorem{prop}{Proposition}

\hypersetup{
colorlinks   = true, %Colours links instead of ugly boxes
urlcolor     = blue, %Colour for external hyperlinks
linkcolor    = blue, %Colour of internal links
citecolor    = blue %Colour of citations
}

\usepackage{titletoc}

\newcommand\DoToC{%
  \startcontents
  \printcontents{}{1}{\textbf{Table of Contents}\vskip3pt\hrule\vskip5pt}
  \vskip3pt\hrule\vskip5pt
}

\newcommand\blfootnote[1]{%
  \begingroup
  \renewcommand\thefootnote{}\footnote{#1}%
  \addtocounter{footnote}{-1}%
  \endgroup
}

\title{An Attention-based Framework for \\ Fair Contrastive Learning}

\author{Stefan K. Nielsen \\
FPT Software \\
AI Center\\
\And
Tan M. Nguyen \\
Department of Mathematics \\
National University of Singapore \\
}

\begin{document}

\maketitle

\begin{abstract}
\blfootnote{Correspondence to: \texttt{stefannvkp@fpt.com}}
Contrastive learning has proven instrumental in learning unbiased representations of data, especially in complex environments characterized by high-cardinality and high-dimensional sensitive information. However, existing approaches within this setting require predefined modelling assumptions of bias-causing interactions that limit the model's ability to learn debiased representations. In this work, we propose a new method for fair contrastive learning that employs an attention mechanism to model bias-causing interactions, enabling the learning of a fairer and semantically richer embedding space. In particular, our attention mechanism avoids bias-causing samples that confound the model and focuses on bias-reducing samples that help learn semantically meaningful representations. We verify the advantages of our method against existing baselines in fair contrastive learning and show that our approach can significantly boost bias removal from learned representations without compromising downstream accuracy.
\vspace{0.2in}

\end{abstract}

\section{Introduction}

Machine learning models are continuing to achieve impressive results across diverse domains. Wider adoption and development of such models pose immense opportunity, yet there simultaneously exists a substantial risk of societal harm in situations where models propagate forward biases encoded in training data~\citep{lv2023duet, creager2019flexibly, madras2018learning}. In particular, existing facial recognition systems demonstrate racial bias in their classifications, failing to recognize people from certain ethnic groups~\citep{cavazos2020accuracy}. In addition, generative language models, such as GPT-2, have been shown to reproduce gender bias in their generated text, for example in systematically assuming doctors are male and nurses are female among other socially biased outcomes~\citep{kirk2021bias, bender2021dangers}.

One effective approach to resolving this problem is fair representation learning \citep{wang2019balanced, khajehnejad2022crosswalk, zhang2023matrix}. This approach recognizes that bias is encoded at the data level and looks to learn representations of the data that preserve relevant semantic content while removing sensitive information related to a specified protected attribute, such as race, gender, age, geographic location, and so on. Specifically, contrastive learning has been used to learn fair representations. This technique learns similar representations for \textit{positively-paired} samples and dissimilar representations for \textit{negatively-paired} samples \citep{chuang2020debiased, tian2020makes, he2020momentum}. {For example, a positive-pair in the vision setting may be augmentations of the same image and a negative-pair may be any pair of distinct images \citep{chen2020simple}. Thus, designing positive and negative pairs in the right way informs the model what features are semantically meaningful and what features are irrelevant in distinguishing samples. This approach then lends itself to fairness when we design positives and negatives such that the model learns representations that are invariant to the protected attribute, thereby removing sensitive information related to the protected attribute from the learned representations.

%~\tanm{You might want to explain what positively-paired and negatively-paired samples are. Folks might not be familiar with these jargons}. \tanm{Thus, designing positive and negative pairs in the right way helps guide the model towards learning semantically relevant features that are invariant to the protected attributes (you might want to explain what protected attributes are since you did not mention it before. Maybe try to use the word protected attributes in the first paragraph and explain it there), thereby, promoting fairness.} 

% Thus, designing positive and negative pairs in the right way helps guide the model towards learning what features it should consider semantically relevant and what features it should be invariant to in distinguishing samples
% Contrastive learning becomes fair when the features we decide the model should be invariant to are protected attributes.

Existing work in fair contrastive learning often assumes the protected attribute to be a binary variable, such as gender or minority status. Popular fair contrastive learning methods include bias-label flipping, bias-label augmenting, and parity-enforcing regularizers \citep{cheng2021fairfil,ling2022learning,zhang2022fairness, shen2021contrastive,barbano2022unbiased, cheng2021fairfil, hong2021unbiased}. These approaches, while effective in the binary setting, are limited in their usability due to their conceptualization of fairness as a binary problem in which samples can only belong to one of two groups in terms of the protected attribute, such as male/female or majority/minority. As a result, they fail to generalize to the harder and more general problem setting of high-cardinality, high-dimensional, and/or continuous protected attributes. Recently, \cite{tsai2022conditional} considers the continuous protected attribute setting and proposes a conditional sampling procedure in which negative pairs are sampled according to their similarity in the bias dimension. This approach, however, requires a pre-defined kernel function which imposes strong assumptions on the bias-causing interactions among samples. This is because the chosen kernel function specifies exactly for any given similarity between negative samples in the bias dimension the relevance of that sample for contrasting with the positive pair. These strong assumptions on the bias-causing interactions among samples limits the model's ability to learn fair representations and additionally requires expensive matrix inversion operations.

% In particular, their main mechanisms for debiasing come from the binary assumption \tanm{how? You need to be more specific here} and 

{\bf Contribution:} We propose the \textbf{Fa}irness-Awa\textbf{re} (FARE) attention mechanism that attends towards bias-reducing samples and avoids bias-causing samples that confound the model. We further leverage sparsification via locality-sensitive hashing~\citep{shrivastava2014asymmetric,andoni2015practical,kitaev2020reformer} to discard extreme bias-causing samples in FARE and propose the Sparse \textbf{Fa}irness-Awa\textbf{re} (SparseFARE) attention. Our approach is based on the assumption that using similar samples in the bias dimension should prevent the protected information from being used to differentiate samples, thereby removing the sensitive information from the learned representations. FARE and SparseFARE are designed to learn a similarity metric across the protected attributes that capture the bias-causing interactions. To train FARE, we derive the new Fair Attention-Contrastive (FAREContrast) loss that expresses the negative samples as the output of the FARE attention mechanism, in which similarity scores of negative samples are conditioned by learned attention scores. Our contribution is three-fold.
\begin{itemize}
    \item We develop FARE, a novel fairness-aware attention mechanism that captures the bias-causing interactions to reduce bias and learn semantically relevant embedding spaces.
    \item We sparsify FARE to enhance its ability to learn fair representation by discarding extreme bias-causing samples, resulting in the SparseFARE attention.
    \item We derive the FAREContrast loss to train FARE.
\end{itemize}
We empirically demonstrate that compared to the baseline methods, FARE alleviates a significantly larger amount of bias without compromising downstream accuracy and with lower computational complexity.

{\bf Notation:} 
Let calligraphic letters represent dataspaces (e.g $\mathcal{X}$), capital letters represent random variables (e.g $X$), lower case letter represent their outcomes (e.g $x$), and $P_\cdot$ represent distributions of the variable in the subscript (e.g $P_X$). We abuse notation slightly and also denote matrices by capital letters and vectors comprising matrices by lower case letters (e.g $Q  = [q_1, \dots, q_n]^\top$ where $Q \in \mathbb{R}^{n \times k}$ and $q_i \in \mathbb{R}^k$), in which cases we make clear that the capital and lower case letters correspond to matrices and vectors rather than random variables and outcomes.

{\bf Organization:} We structure this paper as follows: Section \ref{background} establishes the necessary technical background. Section \ref{fairness meets attention} derives the FARE and SparseFARE attention mechanisms, as well as the FAREContrast objective loss. Section \ref{experiments} provides the empirical validation of our proposed attention-based methods. Section \ref{related work} discusses related work. The paper ends with concluding remarks. Additional details on experimental setup, further results, and other technical details are found in the Appendix.

\section{Background} \label{background}

In this section, we summarize the technical preliminaries needed to develop our method, comprising conditional contrastive learning and attention mechanisms.

\subsection{Conditional Contrastive Learning} 
\label{sec:ccd}

Contrastive methods learn an encoding of the data such that similar samples are near each other while dissimilar samples are far from each other in the 
embedding space \citep{chen2020simple,he2020momentum,hjelm2018learning}. This is done by sampling a positive sample $y_{pos}$ and negative sample $y_{neg}$ for any given $x \in \mathcal{X}$,  where the encoder learns a representation such that $x$ and $y_{pos}$ are near each other while $x$ and $y_{neg}$ are distant. Conditional contrastive methods extend this approach to allow for conditional sampling on an additional variable $Z$, which in the fairness setting is a protected attribute \citep{tsai2022conditional}. In particular, the data pair $(x, y_{pos})$ is sampled from $P_{XY | Z = z}$ as $x$ and $y_{pos}$ are views of one another (obtained via augmentation) and $(x, y_{neg})$ is sampled from $P_{X | Z = z}P_{Y | Z = z}$ as $x$ and $y_{neg}$ are two distinct samples \citep{oord2018representation,tsai2021self}.

% rather than sampling $(x, y_{pos}) \sim P_{XY}$ and $(x, y_{neg}) \sim P_XP_Y$, we condition on $Z$ and sample $(x, y) \sim P_{XY | Z = z}$ or $(x,y) \sim P_{X | Z = z}P_{Y | Z = z}$ depending on the particular problem setup and whether $y$ is a positive or negative sample (\cite{oord2018representation} \cite{tsai2021self}).
% We let $Z$ be the protected attribute and use it for conditionally sampling the negatives. Hence we assume access to bias labels but not class labels. Then, the positive pair is sampled from the joint distribution $P_{XY | Z = z}$ and the negative pair is sampled from the conditional distribution  $y_{neg} \sim P^b_{Y | Z = z}$ where $b$ denotes the batch size.
The Fair-InfoNCE objective (\cite{tsai2021conditional}) is then defined as:
\begin{equation} \label{fair-infoNCE}
    \underset{f}{\text{sup}} \E_{z \sim P_Z, \hspace{0.1cm} (x, y_{pos}) \sim P_{XY | Z = z}, \hspace{0.1cm} \{y_{neg}\}_{i=1}^b \sim P^{\otimes b}_{Y | Z = z}} \left[ \text{log} \frac{e^{f(x, y_{pos})}}{e^{f(x, y_{pos})} + \sum_{i=1}^b e^{f(x, y_{neg, i})}}  \right]
\end{equation} 
where $b$ denotes the batch size and $f: \mathcal{X} \times \mathcal{Y} \rightarrow \mathbb{R}$ is a mapping parameterized by neural networks $g_{\theta_X}, g_{\theta_Y}$, given by:
\begin{align}
\label{eqn:scoring_f}
    f(x, y) =  \text{cosine similarity}\Bigl(g_{\theta_X}(x), g_{\theta_Y}(y)\Bigr)/\tau ,
\end{align}
where the networks are themselves parameterized by $\theta_X, \theta_Y$ and $\tau$ is a hyperparameter scaling the cosine similarity \citep{chen2020simple}. In many cases, as in ours, $g_{\theta_X} = g_{\theta_Y}$. The function $f$ from \ref{eqn:scoring_f} is referred to as the scoring function between samples $x,y$ and evaluates the similarity between the learned embeddings of the neural network. Hence, the learning objective aims to maximize the score for positive pairs and minimize the score for negative pairs.

We also express the exponential scoring function in terms of an inner product in a Reproducing Kernel Hilbert Space (RKHS) with corresponding feature map \citep{tsai2022conditional} as follows:
\begin{align}
    e^{f(x,y)} = \text{exp} \Bigl( \text{cosine similarity}(g_{\theta_X}(x), g_{\theta_Y}(y) )/ \tau \Bigr) := \Bigl\langle \phi(g_{\theta_X}(x) ), \phi ( g_{\theta_Y}(y)) \Bigr\rangle_{\mathcal{H}},
\end{align}
where $\langle \cdot, \cdot \rangle_{\mathcal{H}}$ is the inner product in RKHS $\mathcal{H}$ and $\phi$ is the feature map associated with $\mathcal{H}$.\footnote{Note the exponential of the scoring function is a proper kernel, as the scoring function is a proper kernel and the exponential of proper kernels are proper kernels as well.} 
% We introduce this notation to help derive FARE, which estimates the exponential scoring function by reinterpretting the inner product in the RKHS as an attention computation.}

\subsection{Attention Mechanism} 
\label{sec:attn}
The scaled dot-product attention mechanism (\cite{vaswani2017attention}) is given as:
\begin{align}
\text{Attention}(Q, K, V) = \underbrace{ \text{softmax}\left( \frac{QK^{\top}}{\rho} \right) }_PV, \nonumber
\end{align}
where $Q = TW_Q$, $K = SW_K$ and $V = UW_V$ representing the queries, keys and values respectively, which are obtained via learnable linear projections, $W_Q, W_K \in \mathbb{R}^{d_m \times d_k}, W_V \in \mathbb{R}^{d_m \times d_v} $, of data matrices $S \in \mathbb{R}^{n \times d_m}, T \in \mathbb{R}^{n \times d_m}, U \in \mathbb{R}^{n \times d_m}$ where $n$ is the sequence length, $d_m$ is the embedding dimension and $d_v$ is the chosen hidden dimension of the projection subspaces. The softmax operator is applied row-wise, and $\rho$ is a temperature hyperparameter most often set to $\sqrt{d}$.  We refer to $P \in \mathbb{R}^{n \times n}$ as the attention map, which contains information regarding the learned similarities between individual keys and queries. In many cases, $S = T = U$, referred to as self-attention. Our model is inspired by self-attention, where we take $S = T = Z$, where $Z = [z_1, \dots, z_n]^\top$ is the input sequence of protected attributes, but $U \neq Z$. Instead, for our purposes, we take $U \in \mathbb{R}^{n \times n}$ with entries $[U]_{ij} = e^{f(x_i, y_j)}$ which is the matrix of similarity scores between samples $x_i, y_j$. Furthermore, we pass this matrix straight into the attention computation without projecting it with $W_V$, and so $U = V$ (see Remark 2 in section \ref{fare attention head}). This is because we wish to use the attention map $P$ to provide contextual information to condition the similarity scores, $e^{f(x_i, y_j)}$, rather than the sensitive attributes. Under this setup, the attention score $p_{ij}$ and the output $o_i$ of the attention as follows:
\begin{align}
\label{eqn:attn-mech}
p_{ij} &= \text{softmax}( (W_Q t_i)^{\top}(W_K s_j) / \rho  ), \,\,\, o_i = \sum_j^n p_{ij} e^{f(x_i, y_j)}. 
\end{align}
The output of the attention mechanism can therefore be interpreted as a conditionally weighted sum over the values with weights provided by the attention scores. Section \ref{fairness meets attention} illustrates how these attention scores capture bias-causing interactions, and so the attention outputs are equivalently the values conditioned by their bias-causing potential, which serves the purpose of accentuating bias-reducing samples and attenuating bias-causing samples, which helps to learn debiased representations.

\section{Fairness Meets Attention} \label{fairness meets attention}
In this section, we present our \textbf{Fa}irness-Awa\textbf{re} (FARE) attention mechanism. FARE focuses on negative samples to reduce bias and improve the semantic content of the learned representations. The model passes the negative samples through an attention mechanism where the outputs are the linearly weighted sum of negative samples according to their bias dimension and their semantic relevance, where the weights are the attention scores (see Section~\ref{fare attention head}). The attention matrix is then sparsified such that high bias-inducing samples are given zero attention scores (see Section \ref{sparse}), resulting in the Sparse \textbf{Fa}irness-Awa\textbf{re} attention (SparseFARE). FARE and SparseFARE are trained to minimize a novel Fair Attention-Contrastive (FAREContrast) loss,
which incorporates FARE/SparseFARE into the  Fair-InfoNCE objective in Eqn.~\ref{fair-infoNCE}
\textcolor{black}{FAREContrast loss allows FARE-based methods to capture the bias-causing interactions over samples while learning good representation for downstream tasks.}

% learning to minimize the contrastive loss, thereby focusing attention on samples that are both bias-reducing and that help learn the representation space.} \tanm{Please rephrase this}

% Our method considers a batch of negative samples and focuses attention on negatives that will reduce bias and improve the semantic content of the learned representations. The model does this by passing the negative samples through the \textbf{Fa}irness-Awa\textbf{re} (FARE) attention head where the outputs are the appropriately weighted negative samples according to their bias dimension and their semantic relevance (section \ref{fare attention head}). Within the conditioning attention head, extreme bias-inducing samples are given exactly zero attention scores via sparsification (section \ref{sparse}). The attention head learns in response to the corresponding \textbf{Att}ention-\textbf{Con}trastive (AttCon) loss, and in this way models the bias-interactions and adapts them towards the task of separating out the embedding space (section \ref{criterion}).

\subsection{FARE: Fairness-Aware Attention} \label{fare attention head}
 % FARE is derived from the kernel-based scoring function estimation (\cite{tsai2022conditional}), where we reinterpret this operator as an attention mechanism. The benefit of using attention rather than kernel similarity functions is that rather than heuristically specifying a function to model the bias interactions, FARE learns that function, capturing more flexible bias-interactions and better debiased representations. FARE is also computationally and memory efficient since it does not require the expensive matrix inversion needed in the kernel-based scoring function estimation.

% We start our derivation of FARE with the estimation of the scoring function in Eqn.~\ref{fair-infoNCE}. 
The only available data is the batch of triplets $\{x_i, y_i, z_i\}_{i=1}^{b}$, which are independently sampled from the joint distribution $P^{\otimes b}_{XYZ}$ with $b$ being the batch size, and we do not have access to data pairs from the conditional distribution $P_{X|Z}P_{Y|Z}$. Therefore, we aim to bypass the conditional sampling process from the Fair-InfoNCE objective in Eqn.~\ref{fair-infoNCE}. In particular, to transform the Fair-InfoNCE objective into an alternative version that does not require conditional sampling, we estimate the scoring function $e^{f(x, y)}$ for $ (x,y) \sim P_{X|Z}P_{Y|Z}$ in Eqn.~\ref{fair-infoNCE} given only $\{x_i, y_i, z_i\}_{i=1}^{b} \sim P^{\otimes b}_{XYZ}$. We do this by employing kernel density estimators to view the desired similarity score as the output of an attention mechanism, which leverages attention as kernelized non-linear similiarity score \citep{tsai2019transformer, parzen1962estimation, rosenblatt1956remarks}. \emph{Given an anchor $(x_i, z_i)$, FARE estimates the similarity score between $x_i$ and $y \sim P_{Y|Z = z_i}$ by conditionally weighting all samples in the batch, with weights provided by learned attention scores over the protected attributes.} We derive FARE below.

For any $Z = z$, given $y \sim P_{Y|Z=z}$,  we first follow \cite{tsai2022conditional} and estimate  $\phi(g_{\theta_{Y}}(y))$ with $\mathbb{E}_{y \sim P_{Y|Z=z}} \left[\phi(g_{\theta_{Y}}(y))\right]$, obtaining:
\begin{align}
\phi(g_{\theta_{Y}}(y)) &\approx \mathbb{E}_{y \sim P_{Y|Z=z}} \left[\phi(g_{\theta_{Y}}(y))\right] = \int \phi(g_{\theta_{Y}}(y)) P(y|z) dy = \int \phi(g_{\theta_{Y}}(y)) \frac{P(y,z)}{P(z)} dy \label{eqn:emb_est}.
\end{align}
We then plug Eqn.~\ref{eqn:emb_est} into Eqn.~\ref{eqn:scoring_f} for the data pair $(x_i, z_i)$ to estimate $e^{f(x_i, y)}$ when $y \sim P_{Y|Z=z_i}$ as
\begin{align}
\label{eqn:est-score-1}
\hat{e}_{\text{conditioned}}^{f(x_i, y)} &\approx \Bigl\langle \phi (g_{\theta_{X}}(x_i)), \int \phi(g_{\theta_{Y}}(y)) \frac{P(y,z)}{P(z)} dy \Bigr\rangle_{\mathcal{H}} \nonumber \\
&= \text{tr}\left(\phi (g_{\theta_{X}}(x_i))^{\top} \int \phi(g_{\theta_{Y}}(y)) \frac{P(y,z)}{P(z)} dy\right) \nonumber \\
&= \phi (g_{\theta_{X}}(x_i))^{\top} \int \phi(g_{\theta_{Y}}(y)) \frac{P(y,z)}{P(z)} dy.
\end{align}
Here we denote the conditional estimation of the scoring function $e^{f(x, y)}$ for $ (x,y) \sim P_{X|Z}P_{Y|Z}$ by $\hat{e}_{\text{conditioned}}^{f(x, y)}$.

 {\bf Kernel density estimator.} To estimate $P(y,z)$ and $P(z)$, we employ the kernel density estimation approach~\citep{parzen1962estimation,rosenblatt1956remarks}. In particular, by using the isotropic Gaussian kernel with bandwidth $\sigma$, we obtain the following estimators of $P(y,z)$ and $P(z)$:
\begin{align}
\label{eqn:kde}
    \hat{P}_{\sigma}(y,z) = \frac{1}{b}\sum_{j=1}^{b}\varphi_{\sigma}(y - y_j)\varphi_{\sigma}(z - z_j), \,\,\,\hat{P}_{\sigma}(z) = \frac{1}{b}\sum_{j=1}^{b}\varphi_{\sigma}(z - z_j),  
\end{align}
where $\varphi_{\sigma}(\cdot)$ is the isotropic multivariate Gaussian density function with diagonal covariance matrix $\sigma^{2}\bf{I}$. Given Eqn.~\ref{eqn:est-score-1} and the kernel density estimators in Eqns.~\ref{eqn:kde}, we attain the following conditional estimation of the scoring function:
\begin{align}
\label{eqn:est-score-2}
    \hat{e}_{\text{conditioned}}^{f(x_i, y)} &= \phi (g_{\theta_{X}}(x_i))^{\top} \int \phi(g_{\theta_{Y}}(y)) \frac{\hat{P}(y,z)}{\hat{P}(z)} dy \nonumber \\
    &= \phi (g_{\theta_{X}}(x_i))^{\top} \int \phi(g_{\theta_{Y}}(y)) \frac{\sum_{j=1}^{b}\varphi_{\sigma}(y - y_j)\varphi_{\sigma}(z - z_j)}{\sum_{j=1}^{b}\varphi_{\sigma}(z - z_j)} dy \nonumber \\
    &= \phi (g_{\theta_{X}}(x_i))^{\top} \frac{\sum_{j=1}^{b}\varphi_{\sigma}(z - z_j)\int \phi(g_{\theta_{Y}}(y)) \varphi_{\sigma}(y - y_j) dy}{\sum_{j=1}^{b}\varphi_{\sigma}(z - z_j)} \nonumber \\
    &= \frac{\sum_{j=1}^{b} \left[\phi (g_{\theta_{X}}(x_i))^{\top}\phi(g_{\theta_{Y}}(y_j))\right]\varphi_{\sigma}(z - z_j)}{\sum_{j=1}^{b}\varphi_{\sigma}(z - z_j)}.
\end{align}

{\bf Connection to Attention Mechanism.}  In Eqn.~\ref{eqn:est-score-2}, we replace $\varphi_{\sigma}$ by the formula of the isotropic multivariate Gaussian density function with diagonal covariance matrix $\sigma^{2}\bf{I}$ and obtain
\begin{align}
\label{eqn:est-score-3}
    \hat{e}_{\text{conditioned}}^{f(x_i, y)} &= \frac{\sum_{j=1}^{b} \left[\phi (g_{\theta_{X}}(x_i))^{\top}\phi(g_{\theta_{Y}}(y_j))\right]\exp{\left(-\|z - z_j\|^{2}/2\sigma^{2}\right)}}{\sum_{j=1}^{b}\exp{\left(-\|z - z_j\|^{2}/2\sigma^{2}\right)}} \nonumber \\
    &= \frac{\sum_{j=1}^{b} \left[\phi (g_{\theta_{X}}(x_i))^{\top}\phi(g_{\theta_{Y}}(y_j))\right]\exp{\left(-(\|z\|^{2} + \|z_j\|^{2})/2\sigma^{2}\right)}\exp{\left(z^{\top}z_j/\sigma^{2}\right)}}{\sum_{j=1}^{b}\exp{\left(-(\|z\|^{2} + \|z_j\|^{2})/2\sigma^{2}\right)}\exp{\left(z^{\top}z_j/\sigma^{2}\right)}}
\end{align}
If we further assume that $z_j$, $j=1,2,\dots,b$ are normalized and choose $\sigma^{2}=\rho$, where $\rho$ is the attention temperature hyperparameter in Eqn.~\ref{eqn:attn-mech}, the conditionally estimated scoring function is then
\begin{align}
\label{eqn:est-score-4}
    \hat{e}_{\text{conditioned}}^{f(x_i, y)}
    &= \frac{\sum_{j=1}^{b} \left[\phi (g_{\theta_{X}}(x_i))^{\top}\phi(g_{\theta_{Y}}(y_j))\right]\exp{\left(z^{\top}z_j/\rho\right)}}{\sum_{j=1}^{b}\exp{\left(z^{\top}z_j/\rho\right)}} \nonumber \\
    &= \sum_{j=1}^{b} \text{softmax}\left(z^{\top}z_j/\rho\right) \left[\phi (g_{\theta_{X}}(x_i))^{\top}\phi(g_{\theta_{Y}}(y_j))\right].
\end{align}

% \begin{definition}[Integral Conditional Embedding Operator]
% The Integral 
% \end{definition}

Plugging in the observed outcome of the protected attribute, $z  = z_i$, and allowing $z_i$ and $z_j$ to be transformed by learnable linear transformation, $W_Q, W_K$, the conditionally estimated similarity score $e^{f(x_i, y)}$ when $y \sim P_{Y | Z = z_i}$ is then given by
\begin{align}
\hat{e}_{\text{conditioned}}^{f(x_i, y)} = \sum_{j=1}^{b}\underbrace{\text{softmax}\left((W_Qz_i)^{\top}W_Kz_j/\rho\right)}_{p_{ij}} \underbrace{ \left[\phi (g_{\theta_{X}}(x_i))^{\top}\phi(g_{\theta_{Y}}(y_j))\right]}_{e^{f(x_i, y_j)}},
\end{align}

which is the output of an attention mechanism with values given by the unconditioned similarity scores between samples, $e^{f(x_i, y_j)}$, and attention scores $p_{ij}$ computed over the protected attributes $(z_i, z_j)$. Thus, the similarity scoring function estimation between $x_i$ and $y \sim P_{Y|Z = z_i}$ can be approximated by an attention output. We summarize this new result in the following proposition.

\begin{prop}[Conditional Estimation of $e^{f(x_i, y)}$ when $y \sim P_{Y|Z = z_i}$]
\label{prop:cond_est_score}
Given $\{x_i, y_i, z_i\}_{i=1}^{b} \sim P^{\otimes b}_{XYZ}$, the finite-sample estimation of $e^{f(x_i, y)}$ is $\sum_{j=1}^{b}\text{softmax}\left((W_Qz_i)^{\top}W_Kz_j/\rho\right) \left[\phi (g_{\theta_{X}}(x_i))^{\top}\phi(g_{\theta_{Y}}(y_j))\right]$, which is the output of an attention mechanism.
\end{prop}

Hence, the attention scores $p_{ij}$ condition the similarity scores $e^{f(x_i, y_j)}$, i.e., for any data pair $(x_i, y_j)$, their similarity is accentuated/attenuated depending on the attention between the protected attributes $(z_i, z_j)$. At a high level, when $z_i$ is dissimilar from $z_j$, $x_j$ is likely to cause a bias in the learned representations, and we expect the attention mechanism to divert its focus from that sample. Conversely, when $z_i$ is similar to $z_j$, $x_j$ is likely to reduce the bias in the learned representations, and the attention mechanism should place more focus on that sample. However, rather than specifying the extent to which similarities over the bias dimension should mask out samples via a pre-defined kernel as in~\cite{tsai2022conditional}, we allow the attention mechanism to learn this metric given the task. This flexibility allows the model to focus on samples that are bias-reducing and shift its emphasis away from samples that are bias-causing while simultaneously adapting towards the overall task of learning semantically meaningful representations.

We are now ready to give a full definition of FARE.

\begin{definition}[Fairness-Aware Attention]
\label{def:fare}
    Fairness-aware attention (FARE) is an attention mechanism that computes the finite-sample estimation of the similarity scores $e^{f(x_i, y)}$ when $y \sim P_{Y|Z = z_i}$ for $i=1,2,\dots,b$ with $b$ being the batch size. Given $\{(x_i, y_i, z_i)\}_{i=1}^{b} \sim P^{\otimes b}_{XYZ}$, FARE is defined as
    \begin{align}
    \text{FARE}( \{ (x_i, y_i, z_i) \}_{i=1}^b)= \hat{e}_{\text{conditioned}}^{f(x_i, y)} = \sum_{j=1}^{b}\text{softmax}\left((W_Qz_i)^{\top}W_Kz_j/\rho\right) \left[\phi (g_{\theta_{X}}(x_i))^{\top}\phi(g_{\theta_{Y}}(y_j))\right]
    \end{align}
\end{definition}
FARE estimates the similarity between any given anchor and negative sample, where the similarity is conditioned according to the protected attribute and the extent to which any sample is likely to bias the representations. By focusing attention on samples according to their bias-inducing characteristics, FARE is able to learn fair representations.

\begin{remark}
In Proposition~\ref{prop:cond_est_score}, the attention score $p_{ij} = \text{softmax}\left((W_Qz_i)^{\top}W_Kz_j/\rho\right)$, $i,j = 1,2,\dots,b$, provides a context to estimate the similarity score between $x_i$ and $y \sim P_{Y|Z = z_i}$, thus allowing FARE to attain a contextual representation. It has been shown that the ability of the attention mechanism to capture rich and diverse contextual representation is key to the impressive performance of recent deep learning models, including transformers and graph neural networks~\citep{tenney-etal-2019-bert,vig-belinkov-2019-analyzing,clark-etal-2019-bert,voita-etal-2019-analyzing,hewitt-liang-2019-designing}.
\end{remark}

\begin{remark}
We do not include a learnable value transformation matrix $W_V$ for the values. Rather we pass the unconditioned similarity scores, $e^{f(x_i, y_i)}$, straight into the attention mechanism. This is because a transformation $W_V$ would allow the optimization procedure to take a shortcut and avoid minimizing the objective loss by just sending the value weights to infinity, obtaining 0 loss and thereby preventing the encoder from learning useful representations. More details are given in Appendix \ref{appendix:attcon criterion}.
\end{remark}

% \textbf{Proposed Model.} The full definition of FARE is therefore given as:

% \begin{align}
%     \text{FARE Attention}(Z, U) = \text{softmax}\left(\frac{( ZW_Q)(Z W_K)^\top}{t}\right) {U},
% \end{align}

\subsection{SparseFARE: Sparse Fairness-Aware Attention} \label{sparse}

In the previous section, we proposed the use of attention for debiasing representations, we now discuss the role of sparsification towards this goal. If we have prior knowledge on the proportion of samples that need not be considered at all since they are relatively extreme in the bias dimension, then we can discard those samples before computing attention. For example, if color is the protected attribute, then samples with opposing colors such as black/white may be considered extreme in the bias dimension relative to each other. This allows the attention mechanism to be more efficient and debias more aggressively as samples can be given an attention score of exactly 0. We implement the sparse fairness-aware attention (SparseFARE) via locality-sensitive hashing  (LSH) (\cite{kitaev2020reformer}).

\textbf{Locality-Sensitive Hashing.} A hashing scheme is locality-sensitive if for all vectors, $z$, assigned hashes $h(z)$, similar vectors are assigned the same hash with high probability and dissimilar vectors are assigned the same hash with low probability~\citep{kitaev2020reformer}. We follow the LSH scheme in \citep{andoni2015practical}, which employs random projections $R \in \mathbb{R}^{d_z \times b/2}$ where $[R]_{ij} \sim N(0,1)$ and assigns hashes by $h(z) = \text{argmax}(\text{concat}(zR, -zR))$.

\textbf{Locality-Sensitive Hashing Attention for Fairness.} The basis of the debiasing scheme is the assumption that for anchor $(x_i, z_i)$ and negative sample $(y_j, z_j)$, $y_j$ is likely to increase the bias of the representations when $z_i$ is dissimilar to $z_j$. If we determine some threshold for ignoring $(y_j, z_j)$ when $z_i$ and $z_j$ are sufficiently dissimilar, then we can leverage the LSH scheme to ensure with high probability that $h(y_j) \neq h(x_i)$. Subsequently, if we only permit attention to be calculated within hash buckets (or potentially within hash buckets and across adjacent buckets), we should ignore samples at the relative extremes of $Z$ with high probability to speed up our fairness mechanism and perform more aggressive debiasing by discarding extreme bias-causing samples.

For index $i$ of a given query $q_i$, we denote the attention support as $S_i = \{j: h(k_j) = h(k_i)\}$, which is the set of keys hashed to the same bucket and therefore take part in the attention computation with $q_i$.\footnote{We denote the attention support as purely intra-bucket here for simplicity. In reality, we will typically allow cross-attention to adjacent buckets as well} Figure~\ref{fig:hashing} illustrates this scheme.

\begin{figure}
\centering
\begin{tikzpicture}[scale = 1,
module/.style={draw, thin, minimum width=2ex, minimum height = 2ex},
modulelarge/.style={draw, thin, minimum width=20ex},
smallsampleempty/.style={module},
smallsamplered/.style={module, fill = {rgb:red,100;green,176;blue,255}},
smallsamplered80/.style={module, fill = {rgb:red,100;green,143;blue,255}},
smallsampleyellow/.style={module, fill = {rgb:red,220;green,38;blue,127}},
smallsampleyellow80/.style={module, fill = {rgb:red,220;green,38;blue,127}},
smallsamplecyan/.style={module, fill = {rgb:red,254;green,97;blue,0}},
smallsamplecyan80/.style={module, fill = {rgb:red,254;green,97;blue,0}},
smallsamplegreen/.style={module, fill = {rgb:red,255;green,176;blue,0}},
smallsamplegreen80/.style={module, fill = {rgb:red,255;green,176;blue,0}},
largesampplecyan/.style={module, fill=cyan!40},
largesamplered/.style={module, fill=red!40},
largesampplecyellow/.style={module, fill=yellow!40},
arrow/.style={-stealth', thin, rounded corners, draw = gray!75},
arrowblack/.style={-stealth', thin, rounded corners, draw = gray!99, scale = 0.9}
]

\node[smallsamplered80,label={[align=center,font=\small, label distance = 6mm]left:Sequence of keys/queries:\\$\{z_j\}_{j=1}^b$}, label={[align=center,font=\tiny]below:Anchor}] (b0) {};
\node[right = 6mm of b0, smallsamplered] (b1) {};
\node[right = 1mm of b1, smallsampleyellow] (b2) {};
\node[right = 1mm of b2, smallsampleyellow] (b3) {};
\node[right = 1mm of b3, smallsamplered80] (b4) {};
\node[right = 1mm of b4, smallsamplegreen] (b5) {};
\node[right = 1mm of b5, smallsamplegreen] (b6) {};
\node[right = 1mm of b6, smallsamplecyan] (b7) {};
\node[right = 1mm of b7, smallsamplecyan] (b8) {};
\node[right = 1mm of b8, smallsampleyellow80] (b9) {};
\node[right = 1mm of b9, smallsamplegreen80] (b10) {};
\node[right = 1mm of b10, smallsampleyellow] (b11) {};
\node[right = 1mm of b11, smallsamplered] (b12) {};
\node[right = 1mm of b12, smallsamplecyan] (b13) {};
\node[right = 1mm of b13, smallsamplegreen80] (b14) {};
\node[right = 1mm of b14, smallsamplecyan80] (b15) {};

\node[below = 6mm of b0, label={[align=center,font=\small, label distance = 6mm]left:LSH scheme\\returns buckets of similar $z$}, label={[align=center,font=\tiny]below:Anchor}, smallsamplered80] (ba0) {};
\node[below = 6mm of b1, smallsamplered80] (ba1) {};
\node[below = 6mm of b2, smallsamplered] (ba2) {};
\node[below = 6mm of b3, smallsamplered] (ba3) {};
\node[below = 6mm of b4, smallsampleyellow80] (ba4) {};
\node[below = 6mm of b5, smallsampleyellow] (ba5) {};
\node[below = 6mm of b6, smallsampleyellow] (ba6) {};
\node[below = 6mm of b7, smallsampleyellow] (ba7) {};
\node[below = 6mm of b8, smallsamplecyan] (ba8) {};
\node[below = 6mm of b9, smallsamplecyan] (ba9) {};
\node[below = 6mm of b10, smallsamplecyan] (ba10) {};
\node[below = 6mm of b11, smallsamplecyan80] (ba11) {};
\node[below = 6mm of b12, smallsamplegreen] (ba12) {};
\node[below = 6mm of b13, smallsamplegreen] (ba13) {};
\node[below = 6mm of b14, smallsamplegreen80] (ba14) {};
\node[below = 6mm of b15, smallsamplegreen80] (ba15) {};

\node[below right = 14mm and 0.1mm of ba0, label={[align=center,font=\small, label distance = 11mm]left:Attend within bucket\\and across neighbouring\\buckets}, label={[align=center,font=\tiny]left:Anchor}, smallsamplered80] (bb0) {};
\node[right = 0.5mm of bb0, smallsamplered80] (bb1) {};
\node[right = 0.5mm of bb1, smallsamplered] (bb2) {};
\node[right = 0.5mm of bb2, smallsamplered] (bb3) {};
\node[right = 3mm of bb3, smallsampleyellow80] (bb4) {};
\node[right = 0.5mm of bb4, smallsampleyellow] (bb5) {};
\node[right = 0.5mm of bb5, smallsampleyellow] (bb6) {};
\node[right = 0.5mm of bb6, smallsampleyellow] (bb7) {};
\node[right = 3mm of bb7, smallsamplecyan] (bb8) {};
\node[right = 0.5mm of bb8, smallsamplecyan] (bb9) {};
\node[right = 0.5mm of bb9, smallsamplecyan] (bb10) {};
\node[right = 0.5mm of bb10, smallsamplecyan80] (bb11) {};
\node[right = 3mm of bb11, smallsamplegreen] (bb12) {};
\node[right = 0.5mm of bb12, smallsamplegreen] (bb13) {};
\node[right = 0.5mm of bb13, smallsamplegreen80] (bb14) {};
\node[right = 0.5mm of bb14, smallsamplegreen80] (bb15) {};

\draw[arrow] (b1.south) -- (ba2.north);
\draw[arrow] (b2.south) -- (ba5.north);
\draw[arrow] (b3.south) -- (ba6.north);
\draw[arrow] (b4.south) -- (ba1.north);
\draw[arrow] (b5.south) -- (ba12.north);
\draw[arrow] (b6.south) -- (ba13.north);
\draw[arrow] (b7.south) -- (ba8.north);
\draw[arrow] (b8.south) -- (ba9.north);
\draw[arrow] (b9.south) -- (ba4.north);
\draw[arrow] (b10.south) -- (ba14.north);
\draw[arrow] (b11.south) -- (ba7.north);
\draw[arrow] (b12.south) -- (ba3.north);
\draw[arrow] (b13.south) -- (ba10.north);
\draw[arrow] (b14.south) -- (ba15.north);
\draw[arrow] (b15.south) -- (ba11.north);

\draw[arrowblack] (bb0.north) .. controls ($(bb0.north)+(0,0.2)$) and ($(bb1.north)+(0,0.2)$) .. (bb1.north);
\draw[arrowblack] (bb0.north) .. controls ($(bb0.north)+(0,0.3)$) and ($(bb2.north)+(0,0.3)$) .. (bb2.north);
\draw[arrowblack] (bb0.north) .. controls ($(bb0.north)+(0,0.4)$) and ($(bb3.north)+(0,0.4)$) .. (bb3.north);

\draw[arrowblack] (bb0.south) .. controls ($(bb0.south)+(0,-0.8)$) and ($(bb4.south)+(0,-0.8)$) .. (bb4.south);
\draw[arrowblack] (bb0.south) .. controls ($(bb0.south)+(0,-1)$) and ($(bb5.south)+(0,-1)$) .. (bb5.south);
\draw[arrowblack] (bb0.south) .. controls ($(bb0.south)+(0,-1.2)$) and ($(bb6.south)+(0,-1.2)$) .. (bb6.south);
\draw[arrowblack] (bb0.south) .. controls ($(bb0.south)+(0,-1.4)$) and ($(bb6.south)+(0,-1.4)$) .. (bb7.south);

\draw[arrowblack] (bb0.north) .. controls ($(bb0.north)+(0,0.8)$) and ($(bb8.north)+(0,0.8)$) .. (bb8.north);
\draw[arrowblack] (bb0.north) .. controls ($(bb0.north)+(0,1)$) and ($(bb9.north)+(0,1)$) .. (bb9.north);
\draw[arrowblack] (bb0.north) .. controls ($(bb0.north)+(0,1.2)$) and ($(bb10.north)+(0,1.2)$) .. (bb10.north);
\draw[arrowblack] (bb0.north) .. controls ($(bb0.north)+(0,1.4)$) and ($(bb11.north)+(0,1.4)$) .. (bb11.north);

\draw[decorate,decoration={brace,amplitude=5pt,mirror, raise = 1.5mm}] (bb12.south west) -- (bb15.south east) node[midway,below= 2ex,font=\tiny,align=center] {Discard before\\attention computation};

\end{tikzpicture}
\vspace{-0.15in}
\caption{Sparse Fair-Aware Attention (SparseFARE) using LSH to discard bias-causing samples. Relative to the anchor's protected attribute status (blue), the fairness-aware attention (FARE) first groups the samples according to their bias attribute and discards any samples that are likely to be highly bias-inducing (brown). Attention scores between similar and bias-reducing samples are then computed.}
\label{fig:hashing}
\vspace{-0.15in}
\end{figure}
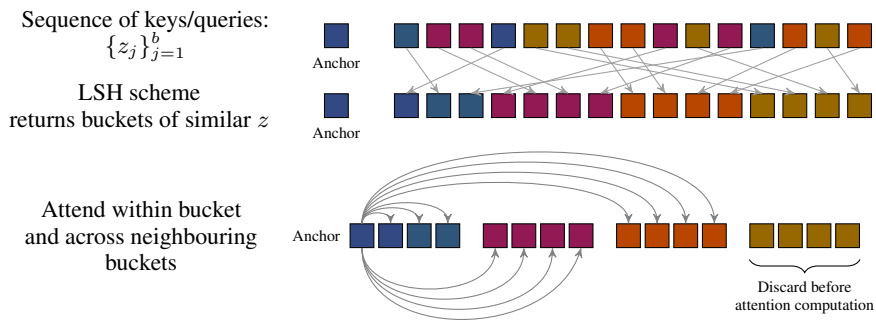

\textbf{SparseFARE Formalization}. Given the LSH scheme for fairness, we now provide the full formulation of the SparseFARE. 
% Denoting the $i^{th}$ output $o_i$:

\begin{definition} [Fairness-Aware Attention with Sparsification]
\label{def:sparsefare}
Sparse fairness-aware attention (SparseFARE) is a variant of FARE in which the attention map over protected attributes is sparsified by removing entries that are highly bias-inducing. Given $\{(x_i, y_i, z_i)\}_{i=1}^b \sim P_{XYZ}^{\otimes b}$, SparseFARE computes the finite-sample estimation of the similarity scores $e^{f(x_i, y)}$ when $y \sim P_{Y | Z = z_i}$ for $i = 1, 2, \dots, b$ with $b$ being the batch size as follows:
\begin{align}
\begin{aligned}
    \text{SparseFARE}( \{(x_i, y_i, z_i) \}_{i=1}^b)&= \hat{e}_{\text{conditioned}}^{f(x_i, y)} \nonumber \\ 
     &\hspace{-12mm}= \sum_{j \in S_i} \text{softmax}\left((W_Qz_i)^{\top}W_Kz_j/\rho - m(j, S_i)\right) \left[\phi (g_{\theta_{X}}(x_i))^{\top}\phi(g_{\theta_{Y}}(y_j))\right], \nonumber
     \end{aligned}
\end{align}
where $S_i = \{ j: h(z_j) = h(z_i)\}$ is the attention support of $i$ and $m(j, S_i) = \begin{cases}
    \infty \hspace{2mm} \text{   if } j \notin S_i\\
    0 \hspace{3mm} \text{   otherwise}
\end{cases}$.  

\end{definition}

\subsection{FAREContrast: Fair Attention-Contrastive Criterion for Contrastive Learning} \label{criterion}
We now present the Fair Attention-Contrastive (FAREContrast) criterion for fair contrastive learning with FARE. We obtain FAREContrast by replacing the summation over negative samples in the Fair-InfoNCE in Eqn.~\ref{fair-infoNCE} with the output of FARE. FAREContrast is then defined as

% \text{FARE}(x_i, y)= \hat{e}_{\text{conditioned}}^{f(x_i, y)} = \sum_{j=1}^{b}\text{softmax}\left((W_Qz_i)^{\top}W_Kz_j/\sigma^{2}\right) \left[\phi (g_{\theta_{X}}(x_i))^{\top}\phi(g_{\theta_{Y}}(y_j))\right]

\begin{equation} \label{eqn:FAREContrast}
    \underset{f}{\text{sup}} \E_{ \{(x_i, y_i, z_i)\}_{i=1}^b \sim P_{XYZ}^{\otimes b}} \left[ \text{log} \frac{e^{f(x_i, y_{i})}}{e^{f(x_i, y_{i})} + \text{FARE}( \{ (x_i, y_{i}, z_i) \}_{i=1}^b )  }  \right].
\end{equation}

The goal of the FAREContrast criterion is to adapt the Fair-InfoNCE objective such that we avoid conditional sampling. We do this because our FARE attention mechanism avoids conditional sampling of negative pairs by using attention to consider the whole batch and selectively weight samples according to their protected attribute status, in this way focusing on bias-reducing samples. Hence we only consider $\{(x_i, y_i, z_i)\}_{i=1}^b \sim P_{XYZ}^{\otimes b}$. Furthermore, we only need FARE for negative samples since only the negatives need to be conditioned for contrasting with the positive pair. The positive pair will necessarily be identical in the bias-dimension as we do not perform augmentations that change the protected attribute. Our method debiases representations by then only showing the positive pair negative samples that have similar protected attribute status, such that the protected information is not used to distinguish samples. Hence, FAREContrast is obtained by replacing the summation over negative samples in Fair-InfoNCE with FARE.

% where $p(\cdot, \cdot)$ denotes the attention score between arguments and $S_i$ is the attention support of $(x_i, z_i)$.

% The attention weights capture the bias-interactions between samples and are then adapted towards optimising the AttCon criterion, thereby tuning the learnt weights towards selecting out the negative samples that best minimize the loss. This allows the FARE attention mechanism to focus attention away bias-causing samples that are confounding the model, helping to learn debiased and semantically meaningful representations.

\section{Experiments} 
\label{experiments}

\begin{table}
\centering
\small
\begin{tabular}{lll} 
\midrule
\textbf{Model} & \textbf{Accuracy ($\uparrow$)} & \textbf{Bias Removal ($\uparrow$)} \\ 
\midrule
\midrule
\multicolumn{3}{c}{\textit{Baseline Models}} \\
\midrule
\midrule
InfoNCE \citep{oord2018representation} & 84.1 $\pm$1.8 & $48.8 \pm 4.5 $ \\
Fair-InfoNCE (\cite{tsai2022conditional}) & $85.9 \pm 0.4$ & $64.9 \pm 5.1$ \\
CCLK (\cite{tsai2022conditional}) & $86.4 \pm 0.9$ & $64.7 \pm 3.9$ \\
\midrule
\midrule
\multicolumn{3}{c}{\textit{Attention-based Models}} \\
\midrule
\midrule
FARE \textbf{(ours)} & $85.7 \pm 0.9$ & $68.4 \pm 4.3$ \\
SparseFARE \textbf{(ours)} & $\textbf{86.4} \pm 1.3$ & $\textbf{74.0} \pm 3.8$ \\
\end{tabular}
\vspace{0.1in}
\caption{Results on colorMNIST. Bias removal is measured by MSE, where high MSE indicates more color information has been removed from the learned representations.}
\label{result:cmnist}
\vspace{-0.15in}
\end{table}

In this section, we numerically justify the advantage of FARE in learning debiased and semantically meaningful representations over the baseline methods including InfoNCE~\citep{oord2018representation}, Fair-InfoNCE~\citep{tsai2022conditional}, SimCLR \citep{chen2020simple} and the conditional contrastive learning with kernel model (CCLK)~\citep{tsai2022conditional}. We aim to show that: (i) our methods are able to learn representations with sensitive information removed, and (ii) our learned representations maintain relevant semantic content. 

\textbf{Datasets.} We conduct our experiments on the ColorMNIST dataset \citep{tsai2022conditional} and CelebA dataset \citep{liu2018large}. ColorMNIST contains 60,000 handwritten digits with a continuous RGB color randomly assigned to the background of each digit. The color is taken to be the protected attribute. CelebA contains 202,599 images of celebrities with 40 binary annotations indicating hair color, gender, and many other attributes. We take Attractive as target and Young and Male as sensitive attributes simultaneously. 

\textbf{Evaluation Protocol.} To evaluate representation quality, we adopt the common technique of freezing the encoder and training a linear classifier using the true labels on the encoded representations and measuring accuracy. To evaluate bias removal in the continuous setting of ColorMNIST, we follow the protocol of \cite{tsai2022conditional} and train a linear layer on the encoded representations to predict each samples' protected attribute. We use the mean squared error (MSE) of predicting the color as a proxy for the extent to which the sensitive information has been removed, where higher loss indicates more sensitive information has been removed. For CelebA, we measure fairness in this binary scenario using the common metric Equalized Odds \citep{hardt2016equality} where a lower score indicates a fairer model. Additional empirical results and experimental details are provided in the Appendix \ref{appendix:exp_details}.

\textbf{Results.} Table \ref{result:cmnist} shows experimental results on the colorMNIST dataset. Our FARE and SparseFARE outperform the baseline methods in terms of bias removal while achieving comparable and better top-1 accuracies. In particular, taking accuracy and bias removal together, SparseFARE is able to weakly Pareto dominate all comparative models, learning substantially less biased representations without compromising downstream accuracy.

Table \ref{result:celeba} shows the results of the attention-based models on the CelebA dataset.  SimCLR achieves highest accuracy while SparseFARE achieves the best fairness. Given that Young and Male are both highly correlated with Attractive, it is intuitive that SimCLR attains top accuracy, as SimCLR does not attempt to remove information relating to these two attributes and so is able to leverage the correlation between attributes and target to make more accurate predictions. SparseFARE Pareto dominates all kernel models in terms of fairness and accuracy except for Linear and Polynomial, which achieve marginally higher accuracy. SparseFARE nonetheless attains a better fairness-accuracy tradeoff curve than these two kernels and so for any given level of accuracy, SparseFARE obtains fairer results (see Appendix \ref{appendix: additional results}).

{\bf Efficiency Analysis.} Our methods are more efficient than the kernel-based baselines. CCLK, by requiring matrix inversion, costs $O(b^3)$, while FARE costs $O(b^2)$ and SparseFARE costs $O(b \text{ log}(b))$ (see Appendix \ref{prop:kernel scoring}). 
% We do not include empirical results on speed, however, as the computational speedup would be observed at high batch sizes, for which we encountered RAM limitations. Hence we note only the theoretical complexity benefits of our models in avoiding matrix inversion and by using fainess-motivated sparsification.

\section{Related Work} \label{related work}

\begin{wraptable}{r}{0.45\textwidth}
\centering
\small
\vspace{-0.1in}
\begin{tabular}{lll} 
\midrule
\textbf{Model} & \textbf{Acc. ($\uparrow$)} & \textbf{EO ($\downarrow$)} \\ 
\midrule
SimCLR \citep{chen2020simple} & \textbf{77.7} & 39.6 \\
\midrule
\midrule
\multicolumn{3}{c}{\textit{Kernel-based Models \citep{tsai2022conditional}}} \\
\midrule
\midrule
CCLK-Cosine & 70.2 & 22.4 \\
CCLK-RBF & 69.9 & 21.8 \\
CCLK-Linear  & 71.1 & 21.1 \\
CCLK-Polynomial  & 71.0 & 20.8 \\
CCLK-Laplacian  & 70.0 & 20.8 \\
\midrule
\midrule
\multicolumn{3}{c}{\textit{Attention-based Models}} \\
\midrule
\midrule
FARE (\textbf{ours}) & 73.7 & 23.5 \\
SparseFARE (\textbf{ours}) & 70.4 & \textbf{18.7} \\
\end{tabular}
\caption{CelebA Results. Fare and SparseFARE in comparison with kernel baselines under various kernel specifications.}
\vspace{-0.2in}
\label{result:celeba}
\end{wraptable}The majority of the literature on fair contrastive learning has considered only binary protected attributes (\cite{park2022fair, chai2022self}). With binary protected attributes, debiasing can be achieved by forming positive pairs as samples with opposing bias classes (\cite{cheng2021fairfil, hong2021unbiased, shen2021contrastive}). Another approach is to form positive pairs by using auxiliary models to learn optimal augmentations that obfuscate the bias class of the sample (\cite{ling2022learning, zhang2022fairness}). This paper proposes an attention-based framework to deal with more general notions of fairness that accommodate high cardinality or continuous protected attributes, whereby we learn semantically meaningful representations such that the protected information has been removed. \cite{tsai2022conditional} also consider this setting and use kernel similarity functions to weigh negative samples along the bias dimension for contrastive learning. Our approach differs from their method by using an attention mechanism to learn the bias-causing interactions among samples without specifying a pre-defined kernel.

Our paper also connects to the growing literature surrounding kernel and attention. Most existing work has looked at decomposing the attention computation and enriching or explaining this mechanism by interpreting it as a kernel function. \cite{tsai2019transformer} propose novel attention mechanisms based on differing kernel functions and \cite{song2021implicit} propose enriching attention with implicit kernel estimation, while \cite{tao2023nonlinear} explain attention through nonlinear SVD of asymmetric kernels and \cite{wright2021transformers} view attention as infinite-dimensional non-mercer binary kernel machines. In contrast, our work derives an attention mechanism from a kernel-based method to learn a task-specific similarity metric that can capture the bias-interaction structure and assist the training procedure to learn better-debiased representations. 

Lastly, sparse attention has been studied in the context of efficient transformers. Sparsity in attention mechanisms has been implemented via sparse factorization (\cite{child1904generating}), via local windows (\cite{beltagy2020longformer}), and via locality-sensitive hashing (\cite{kitaev2020reformer}). While our work leverages locality-sensitive hashing, it does not do so merely to save on computational costs. Rather, locality-sensitive hashing supplements the debiasing scheme by sparsifying the entries of the attention map corresponding to extreme bias-inducing samples. To the best of our knowledge, ours is among the early works of using locality-sensitive hashing, or sparsification in general, for learning fair representations.

\section{Concluding Remarks}

In this paper, we present the Fairness-Aware (FARE) attention mechanism, the Sparse Fairness-Aware (SparseFARE) attention mechanism, and the corresponding Fair Attention-Contrastive (FareContrast) criterion for learning fair representations. We address the difficult problem setting of high cardinality or continuous protected attributes and show that FARE and SparseFARE are able to learn a similarity metric over protected attributes that captures the bias-causing interactions among samples, while also focusing on bias-causing samples that are confounding the model. As a result, our attention-based approach is able to learn debiased and semantically meaningful representations. A limitation of our method is that they only capture one attention pattern between protected attributes, thereby providing only one single context to condition the similarity scores. It is indeed necessary to extend FARE and SparseFARE to a multi-head attention setting to capture more diverse contextual representations. We leave this interesting research direction as future work.

% is that we rely solely on the attention mechanism to focus on bias-reducing negative samples to contrast with the positive pair. One interesting direction for future research would be to attempt to directly regularize the embedding space to try enforce fairness in the representations more directly.

% Our approach avoids the issues of earlier methods, in particular we do not require heuristically specifying any given kernel similarity functions or expensive matrix inversion. Furthermore, in this paper we proposed the first fairness-motivated attention sparsification, which we demonstrate can more aggressively debias the learnt representations without compromising their semantic quality.

\newpage

\bibliography{bib}
\bibliographystyle{bib}

\newpage
\appendix
\begin{center}
{\bf \Large Appendix for ``Fairness-Aware Attention for Contrastive Learning''}
\end{center}

\DoToC

\begin{appendices}

\section{Experimental Details}
\label{appendix:exp_details}
This section provides the details of the model and training for experiments in Section~\ref{experiments}. 

\subsection{Training and Evaluation} 

\textbf{ColorMNIST.} Samples in the colorMNIST dataset are 32x32 resolution handwritten digit images, where the digit is represented in black and the background is some known assigned color which is representable as a continuous RGB color vector. The train-test split is 60,000 training images to 10,000 test images. The augmentation scheme is randomized resized crop followed by a random horizontal flip. We pre-train using the LARS optimizer (\cite{you2017large}) and cosine annealing for the learning rate scheduler. The full FARE attention mechanism with sparsification uses 8 rounds of hashing, a bucket size of 64, and backwards and forwards cross-bucket attention. The linear classifier is trained using L-BFGS as optimizer over 500 iterations. We pre-train with a batch size of 256 for 50 epochs.

\begin{figure}[H]
    \centering
    \includegraphics[scale = 0.4]{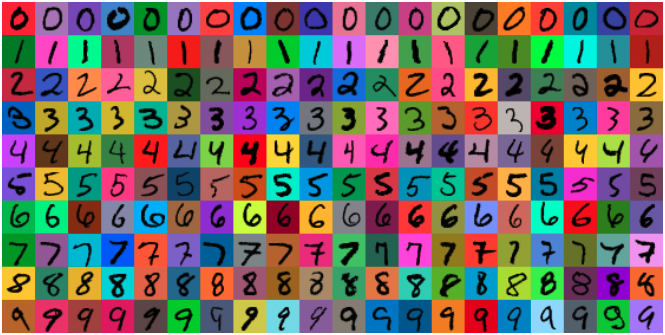}
    \caption{colorMNIST dataset \citep{tsai2022conditional}}
    \label{fig:enter-label}
\end{figure}

We follow the recent contrastive learning literature (\cite{chen2020simple}, \cite{robinson2020contrastive}, \cite{wu2020conditional}) and pre-train the full model before discarding everything except the backbone encoder at evaluation time.

\textbf{CelebA.} The train-test split is the default as provided by PyTorch. Images are resized to 128$\times$128. Resnet-18 \citep{he2016deep} is the encoder and we use the same 2-layer MLP and random augmentation strategies as \cite{chen2020simple}. Same as with colorMNIST, we pre-train with the LARS optimizer and use cosine annealing. We use a batch size of 512 and the LSH scheme uses buckets of size 128 with 8 rounds of hashing and backwards and forwards cross-bucket attention. We train the full model for 100 epochs and evaluate with a single linear layer trained on the frozen encodings for 10 epochs using Adam as optimizer. 

To evaluate the fairness of the representations, we adopt the Equalized Odds (EO) metric \citep{hardt2016equality}. Following \cite{jung2022learning} and \cite{zhang2022fairness}, we compute the metric over multiple sensitive attributes by:

\begin{equation} \label{EO}
    \underset{\forall s^i, s^j \in S}{\text{max}} \overline{\sum}_{\forall y, \hat{y}} \bigg| P_{s^i} \left( \hat{Y} = \hat{y} | Y = y \right) -  P_{s^j} \left( \hat{Y} = \hat{y} | Y = y \right) \bigg|,
\end{equation} 

where $\overline{\sum}$ is the averaged sum, $Y$ is the target label, $\hat{Y}$ is the predicted label, and $s_i, s_j \in S$ are values of sensitive attributes. A smaller EO means a fairer model.

\subsection{Baselines}  

\textbf{ColorMNIST.} The relevant baselines for comparison are the InfoNCE model (\textbf{InfoNCE}) \citep{oord2018representation}, the Fair-InfoNCE model with clustering (\textbf{Fair-InfoNCE}) \citep{tsai2022conditional} and the conditional contrastive learning with kernel model (\textbf{CCLK}) (\cite{tsai2022conditional}).

The InfoNCE model uses the InfoNCE loss function ~\ref{infonce} without performing any conditional sampling. The Fair-InfoNCE model uses the Fair-InfoNCE loss function ~\ref{fair-infoNCE} and performs conditional sampling by first clustering the protected attribute so as to discretize it and then sampling from within the same cluster as the anchor. We report this model's results according to its best performing cluster size as determined by its authors, which is found to be a 10-cluster partition. CCLK uses a kernel similarity metric for weighing negative samples in the batch according to their similarity in the bias-dimension. We report its results according to its best performing kernel choice as chosen by its authors which was the cosine kernel.

The InfoNCE objective \citep{oord2018representation} used in the baseline model InfoNCE is given by:

\begin{equation} \label{infonce}
    \underset{f}{\text{sup}} \E_{(x, y_{pos}) \sim P_{XY}, \hspace{0.1cm} \{y_{neg}\}^n_{i=1} \sim P^{\otimes n}_Y} \left[ \text{log} \frac{e^{f(x, y_{pos})}}{e^{f(x, y_{pos})} + \sum_{i=1}^b e^{f(x, y_{neg, i})}}  \right]
\end{equation} 

\textbf{CelebA.} We compare with \textbf{SimCLR} \citep{chen2020simple} and all kernel implementations of CCLK provided by \cite{tsai2022conditional}. For each kernel model, the kernel in the name refers to the what kernel similarity metric is chosen for measuring the similarity across protected attributes, which then determines the relevance of that sample for being contrasted with the positive sample. For example, CCLK-RBF uses the RBF kernel to compute similarity between two protected attributes.

\section{Connection between Kernel-based Scoring Function Estimation in~\citep{tsai2022conditional} and Attention} \label{prop:kernel scoring}

The CCLK model uses the following kernel-based scoring function estimation:

\begin{prop} [Kernel-Based Scoring Function Estimation \citep{tsai2022conditional}]
\label{tsai estimator}
Given $\{x_i, y_i, z_i\}^b_{i=1} \sim P^b_{XYZ}$, the similarity score of the data pair $(x_i, y)$ given the anchor $z_i$ is computed via the finite-sample kernel estimation $e^{f(x_i, y)}$ when $y \sim P_{Y | Z = z_i}$ as follows:
\begin{align}
    e^{f(x_i, y)} = \left[K_{XY}(K_Z+\lambda \textbf{I})^{-1}K_Z\right]_{ii}, 
\end{align}
\end{prop}

for $i=1, \dots, b$, $[K_{XY}]_{ij} := e^{f(x_i, y_j)}$, and $[K_Z]_{ij} := \langle \gamma(z_i), \gamma(z_j) \rangle_\mathcal{G}$, where $\gamma $ is some kernel feature embedding, $\mathcal{G}$ is the corresponding Reproducing Kernel Hilbert Space (RKHS), and $\langle \cdot, \cdot \rangle_{\mathcal{G}}$ is an inner product in space $\mathcal{G}$.

First, in comparison to Eqn. \ref{tsai estimator}, FARE and sparseFARE avoid matrix inversion. FARE's attention computation has complexity $O(b^2)$ \citep{vaswani2017attention} and sparseFARE has complexity $O(b \hspace{2mm} log b)$ \citep{kitaev2020reformer}, which improve significantly over $O(b^3)$ in Eqn. \ref{tsai estimator}. 

Second, our methods do not impose assumptions on the bias-causing interactions over protected attributes. In particular, we avoid specifying any particular kernel and allow our attention mechanism to learn the bias-causing interactions. To see this difference, we decompose the estimator in Eqn. \ref{tsai estimator} as follows:

\begin{align}
    &\left[K_{XY}(K_Z+\lambda \textbf{I})^{-1}K_Z\right]_{ii} \\
    &= [K_{XY}]_{i*}[(K_Z+\lambda \textbf{I})^{-1}K_Z]_{*i} \nonumber \\
    &= \sum_j^b w(z_i, z_j)e^{f(x_i, y_j)},
\end{align}

where $w(z_i, z_j) =  [(K_Z + \lambda I)^{-1}K_Z]_{ij}$ are smoothed kernel similarity scores \citep{tsai2022conditional}. Hence we see the \citep{tsai2022conditional} estimator as performing a similar weighting of similarity scores between samples, with weights provided by the similarities over the protected attributes. This approach differs from ours however since the kernel must be pre-specified in $K_Z$. This imposes strong assumptions on bias-causing interactions that limit the extent to which the model can learn fair representations. Our method by contrast can be understood as replacing $w(z_i, z_j)$ with attention score $p(z_i, z_j)$. The attention mechanism can more flexibly model the bias-causing interactions and learns to focus-attention on bias-reducing samples that help learn the representation space.

We provide a proof adapted from \citep{tsai2022conditional} of their kernel-based scoring function estimation below.
\begin{proof} [Proof of kernel-based scoring function estimation]

First, letting $\Phi = [\phi(g(y_1)), \dots \phi(g(y_b))]^{\top}$ be the  matrix of kernel embeddings for encodings $g(y_i)$ with feature map $\phi$ and $\Gamma = [\gamma(z_1), \dots , \gamma(z_b)]^\top$ be the matrix of kernel embeddings for protected attribute outcomes $z$ with feature map $\gamma$,  Definition~\ref{song def}  provides the Kernel Conditional Embedding Operator \citep{song2013kernel}:

\begin{definition} \label{song def} [Kernel Conditional Embedding Operator \citep{song2013kernel}]
The finite-sample kernel estimation of $\E_{y \sim P_{Y | Z = z}}\big[\phi(g(y))\big] $ is $\Phi^\top(K_Z + \lambda \textbf{I})^{-1}\Gamma \gamma(z)$ where $\lambda$ is a hyperparameter.
\end{definition}

Then, according to Definition \ref{song def}, for any given $Z = z$, $\phi(g(y))$ when $y \sim P_{Y | Z = z}$ can be estimated by 
\begin{equation}
    \Phi^\top(K_Z + \lambda \textbf{I})^{-1}\Gamma \gamma(z)
\end{equation} 

We look for the inner product between (5) and the encoding of $(x_i, z_i)$ when $y \sim P_{Y | Z = z_i}$:
\begin{align} \nonumber
    &\langle \phi(g(x_i)), \Phi^\top(K_Z + \lambda \textbf{I})^{-1}\Gamma \gamma(z_i) \rangle_{\mathcal{H}} = tr\Bigl( \phi(g(x_i)\Bigr)^\top \Phi^\top (K_Z + \lambda \textbf{I})^{-1} \Gamma \gamma(z_i) \\
    &= [K_{XY}]_{i*}(K_Z + \lambda \textbf{I})^{-1}[K_Z]_{i*} = [K_{XY}]_{i*}[(K_Z + \lambda \textbf{I})^{-1}K_Z]_{*i} \nonumber \\
    &= [K_{XY}(K_Z + \lambda \textbf{I})^{-1}K_Z]_{ii} 
\end{align}  
\end{proof}
% \section{Dataset and Training} \label{appendix:dataset and training}

\section{Comparison of Fair-InfoNCE and FAREContrast} \label{appendix: fairinfonce vs farecontrast}

We present a discussion of the differences between the Fair-InfoNCE objective from \cite{tsai2021conditional} and the FAREContrast objective we use to train our attention-based FARE models. FAREContrast is derived from Fair-InfoNCE by replacing the conditionally sampled negative pairs with the output of the FARE attention mechanism. This leads to a difference firstly in sampling procedure and secondly in the inclusion of learnable attention scores in the loss.

The Fair-InfoNCE \citep{tsai2021conditional} is given as:
\begin{equation} \label{fair-infoNCE}
    \underset{f}{\text{sup}} \E_{z \sim P_Z, \hspace{0.1cm} (x, y_{pos}) \sim P_{XY | Z = z}, \hspace{0.1cm} \{y_{neg}\}_{i=1}^b \sim P^{\otimes b}_{Y | Z = z}} \left[ \text{log} \frac{e^{f(x, y_{pos})}}{e^{f(x, y_{pos})} + \sum_{i=1}^b e^{f(x, y_{neg, i})}}  \right],
\end{equation} 
and FAREContrast is given as:
\begin{equation} \label{eqn:FAREContrast}
    \underset{f}{\text{sup}} \E_{ \{(x_i, y_i, z_i)\}_{i=1}^b \sim P_{XYZ}^{\otimes b}} \left[ \text{log} \frac{e^{f(x_i, y_{i})}}{e^{f(x_i, y_{i})} + \sum_{j=1}^{b}\text{softmax}\left((W_Qz_i)^{\top}W_Kz_j/\rho\right) e^{f(x_i, y_j)}  }  \right],
\end{equation}

where $b$ denotes the batch size, $f: \mathcal{X} \times \mathcal{Y} \rightarrow \mathbb{R}$ is a mapping given by $f(x, y) =  \text{cosine similarity}\Bigl(g_{\theta_X}(x), g_{\theta_Y}(y)\Bigr)/\tau$, $g_{\theta_X}, g_{\theta_Y}$ are neural networks parameterized by $\theta_X, \theta_Y$, and $\tau$ is a hyperparameter scaling the cosine similarity.

We see that FAREContrast does not require conditional sampling of the negatively paired samples, $\{y_{neg}\}_{i=1}^b \sim P^{\otimes b}_{Y | Z = z}$ for outcome of the of the protected attribute $z$. Instead, FARE considers the whole batch and selectively weights samples according to their protected attribute status. One issue with conditional sampling as in Eqn. \ref{fair-infoNCE} is data scarcity, whereby conditioning on $Z = z$ can lead to insufficient negative samples for contrasting \citep{tsai2022conditional}. This problem is exacerbated when the protected attribute has high cardinality or is continuous, which is the problem setting we aim to deal with. When there are insufficient negative samples, we incur risk of poorly learnt representations and collapse \citep{chen2020simple, chen2021exploring}. For this reason, we derive FARE which considers the whole batch and uses learnt attention scores to accentuate/attenuate negative samples according to their bias characteristics. 

The second difference is then the attention weights included in FAREContrast. Including the attention weights in FAREContrast means that that FARE learns according to information coming from the gradients and so can better focus on samples that help minimize the loss, thereby helping the encoder to learn meaningful representations.

\section{Additional Results} \label{appendix: additional results}
\subsection{LSH Bucket Scheme}

\begin{table}[t] 
\centering
\small
\begin{tabular}{lll}
\midrule
\textbf{Attention Scheme} & \textbf{Top-1 Test Accuracy ($\uparrow$)} & \textbf{Bias Removal ($\uparrow$)} \\ 
\midrule
\midrule
Adjacent & $86.4 \pm 1.3$ & $74.0 \pm 3.8$ \\
Intra & $ 84.9 \pm 2.1$ & $ 58.2 \pm 9.8$ \\
\end{tabular}
\vspace{0.0in}
\caption{Sparsification Scheme on ColorMNIST Results. Bias removal is measured by MSE, where high MSE indicates more color information has been removed from the learned representations.}
\label{results:bucket}
\vspace{-0.2in}
\end{table}

Table \ref{results:bucket} shows results for when the LSH scheme considers intra-bucket attention versus the standard adjacent bucket attention (where attention is computed across adjacent buckets). We see fairly substantial drop in performance when restricting attention to within the same bucket, both in terms of accuracy and fairness. Lower accuracy is intuitive given the intra-bucket attention removes three quarters of negative samples, which depletes the model's ability to learn meaningful representations. At the same time, we see lower fairness, despite the heavy debiasing scheme. This may support the conclusion that to learn effectively debiased representations, the model needs sufficiently many samples to learn to attend over and focus on bias-reducing samples. With too few samples in the batch, the model is ignoring too many samples, including ones that would help it learn debiased representations.

\subsection{Fairness-Accuracy Tradeoff}

The two metrics that capture both representation quality and fairness are Accuracy and Equalized Odds (EO). Table \ref{result:celeba} showed that SparseFARE Pareto dominates all kernel baselines in terms of both fairness and accuracy, with the exception of CCLK-Linear and CCLK-Polynomial, which were able to attain slightly higher accuracy. We therefore further compare SparseFARE to these two models by plotting the fairness-accuracy tradeoff curves in Figure \ref{fig:fairness-accuracy}. The curves are produced by plotting EO and Accuracy at four stages during training - after 25, 50, 75, and 100 epochs. We see that for every level of accuracy, SparseFARE achieves better fairness (lower EO). This implies that SparseFARE attains a better fairness-accuracy tradeoff. Additionally of interest, we find that SparseFARE is even able to simultaneously minimize EO while increasing accuracy, implying its accuracy gains are stemming from better quality representations for protected groups.

\subsection{Comparison With Work in Partial Access to Sensitive Attributes}

\begin{table}[H]
\centering
\small
\begin{tabular}{lll} 
\midrule
\textbf{Model} & \textbf{Test Accuracy ($\uparrow$)} & \textbf{EO ($\downarrow$)} \\ 
\midrule
\midrule
\multicolumn{3}{c}{\textit{Supervised Models}} \\
\midrule
\midrule
CGL + G-DRO \citep{sagawa2019distributionally} & 71.4 & 21.9 \\
CGL + FSCL \citep{park2022fair} & 74.0 & 25.6 \\
\midrule
\midrule
\multicolumn{3}{c}{\textit{Unsupervised Models}} \\
\midrule
\midrule
CGL + VFAE \citep{louizos2015variational} & 72.7 & 28.7 \\
CGL + GRL \citep{raff2018gradient} & 73.8 & 26.9 \\
SimCLR \citep{chen2020simple} & \textbf{77.7} & 39.6 \\
FairCL \citep{zhang2022fairness} & 74.1 & 24.5 \\
\midrule
FARE (\textbf{ours}) & 73.7 & 23.5 \\
SparseFARE (\textbf{ours}) & 70.4 & \textbf{18.7} \\
\end{tabular}
\caption{CelebA Results. Fare and SparseFARE in comparison with unsupervised and supervised models under partial sensitive label access.}
\label{results: zhang}
\end{table}

This paper uses the same experimental setup on CelebA as \cite{zhang2022fairness} in terms of training procedure and evaluation protocol. \cite{zhang2022fairness} differs, however, in the sense that the authors assumes only partial access to sensitive attributes and therefore use auxiliary models, for example an editor \citep{zhang2022fairness} or CGL \citep{jung2022learning}, to solve this problem. Given the experimental setups are the same, we include their results as well for reference, however we do not feature these results in the main body given the important difference regarding sensitive attribute access.

\section{Fair Attention-Contrastive Criterion} \label{appendix:attcon criterion}

We do not include a learnable value transformation $W_V$ on the raw similarity scores such that $V = UW_V$ where $U = [e^{f(x_i, y_j)}]_{ij}$ as doing so allows the optimization process to obtain 0 loss without learning meaningful representations. This is seen immediately from the criterion, where allowing $W_V$ gives individual similarity scores as $w_{ij}e^{f(x_i, y_j)}$ in the criterion:

$$ \underset{f}{\text{sup}} \E_{ \{(x_i, y_i, z_i)\}_{i=1}^b \sim P_{XYZ}^b} \left[ \text{log} \frac{e^{f(x_i, y_i)}}{e^{f(x_i, y_i)} + \underset{j \in S_i}{\sum} p(z_i, z_j)w_{ij}e^{f(x_i, y_j)}  }  \right] $$

hence the loss is minimised by sending $w_{ij} \rightarrow \infty \hspace{2mm} \forall i,j$.

\section{Ethical Considerations} \label{appendix: ethics}
\begin{wrapfigure}{R}{0.4\textwidth}
    \centering
    \small
    \includegraphics[scale = 0.28]{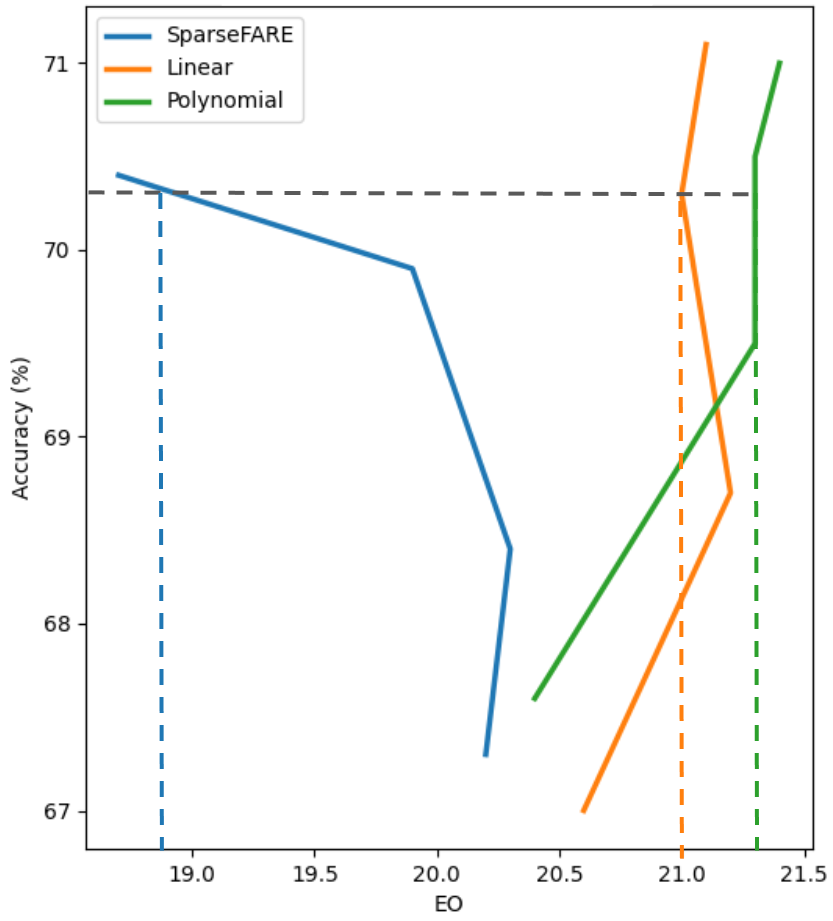}
    \caption{Fairness-Accuracy Tradeoff of SparseFARE and CCLK-}
    \label{fig:fairness-accuracy}
\end{wrapfigure} We note that there are two, interconnected prevalent ethical issues in fair ML. The first is that almost all fair ML literature simplifies the problem of fairness to simple binaries and the second is that fairness metrics (which are typically built atop these binaries) and the choice of which to use themselves involve value judgements that can disadvantage certain people. People have intersectional identities and invariably belong to multiple groups simultaneously. When it comes to choosing fairness metrics,  inherent to the majority of approaches in fair ML is that the researcher or practitioner decide what definition of fairness to use for their model. It has been shown that various definitions of fairness are not only mutually inconsistent but also prioritise different groups in different scenarios \citep{garg2020fairness}. In a sense then, solving for fairer ML models only pushes the problem from the model and onto the practitioner, as a ‘fairer’ model itself advantages and disadvantages different groups under different settings.

These two ethical considerations motivate the approach of our paper to conceptualise fairness in a more general setting where sensitive attributes can be continuous and multi-dimensional and fairer models are measured in terms of sensitive information removal. This conception avoids the ethical issues of binaries and fairness metrics.

We do note however that there still exist ethical concerns with our approach in terms of explainability. Measuring fairness by sensitive information removal (by measuring loss from a trained classifier) does not have an intuitive scale or unit of measurement for discussing the fairness or unfairness of a model. Although we can compare two models in terms of which is fairer, saying a model is fair because it scores some number in MSE has little intuitive meaning. Being unable to communicate the specifics of how a learned representation has removed sensitive information and how will affect downstream classifiers risks undermining confidence in fair ML as well. 

Despite the explainability issue, we nonetheless believe that this approach represents a promising and exciting direction in fair ML that deal with substantive existing ethical issues. We hope that one area of future research may be deriving theoretical frameworks that can derive guarantees between sensitive information removal from debiased representations and upper bounds on downstream fairness metrics. This would develop a practical link to well-known ideas of fairness and how unfair outcomes could appear in worst-case scenarios.

\end{appendices}

\end{document}

%% file: math_commands.tex
\usepackage{amsmath,amsfonts,bm}

\def\eqref#1{equation~\ref{#1}}
\def\1{\bm{1}}

\DeclareMathAlphabet{\mathsfit}{\encodingdefault}{\sfdefault}{m}{sl}
\SetMathAlphabet{\mathsfit}{bold}{\encodingdefault}{\sfdefault}{bx}{n}

%% file: main.bbl
\begin{thebibliography}{56}
\providecommand{\natexlab}[1]{#1}
\providecommand{\url}[1]{\texttt{#1}}
\expandafter\ifx\csname urlstyle\endcsname\relax
  \providecommand{\doi}[1]{doi: #1}\else
  \providecommand{\doi}{doi: \begingroup \urlstyle{rm}\Url}\fi

\bibitem[Andoni et~al.(2015)Andoni, Indyk, Laarhoven, Razenshteyn, and Schmidt]{andoni2015practical}
Alexandr Andoni, Piotr Indyk, Thijs Laarhoven, Ilya Razenshteyn, and Ludwig Schmidt.
\newblock Practical and optimal lsh for angular distance.
\newblock \emph{Advances in neural information processing systems}, 28, 2015.

\bibitem[Barbano et~al.(2022)Barbano, Dufumier, Tartaglione, Grangetto, and Gori]{barbano2022unbiased}
Carlo~Alberto Barbano, Benoit Dufumier, Enzo Tartaglione, Marco Grangetto, and Pietro Gori.
\newblock Unbiased supervised contrastive learning.
\newblock \emph{arXiv preprint arXiv:2211.05568}, 2022.

\bibitem[Beltagy et~al.(2020)Beltagy, Peters, and Cohan]{beltagy2020longformer}
Iz~Beltagy, Matthew~E Peters, and Arman Cohan.
\newblock Longformer: The long-document transformer.
\newblock \emph{arXiv preprint arXiv:2004.05150}, 2020.

\bibitem[Bender et~al.(2021)Bender, Gebru, McMillan-Major, and Shmitchell]{bender2021dangers}
Emily~M Bender, Timnit Gebru, Angelina McMillan-Major, and Shmargaret Shmitchell.
\newblock On the dangers of stochastic parrots: Can language models be too big?
\newblock In \emph{Proceedings of the 2021 ACM conference on fairness, accountability, and transparency}, pp.\  610--623, 2021.

\bibitem[Cavazos et~al.(2020)Cavazos, Phillips, Castillo, and O’Toole]{cavazos2020accuracy}
Jacqueline~G Cavazos, P~Jonathon Phillips, Carlos~D Castillo, and Alice~J O’Toole.
\newblock Accuracy comparison across face recognition algorithms: Where are we on measuring race bias?
\newblock \emph{IEEE transactions on biometrics, behavior, and identity science}, 3\penalty0 (1):\penalty0 101--111, 2020.

\bibitem[Chai \& Wang(2022)Chai and Wang]{chai2022self}
Junyi Chai and Xiaoqian Wang.
\newblock Self-supervised fair representation learning without demographics.
\newblock \emph{Advances in Neural Information Processing Systems}, 35:\penalty0 27100--27113, 2022.

\bibitem[Chen et~al.(2020)Chen, Kornblith, Norouzi, and Hinton]{chen2020simple}
Ting Chen, Simon Kornblith, Mohammad Norouzi, and Geoffrey Hinton.
\newblock A simple framework for contrastive learning of visual representations.
\newblock In \emph{International conference on machine learning}, pp.\  1597--1607. PMLR, 2020.

\bibitem[Chen \& He(2021)Chen and He]{chen2021exploring}
Xinlei Chen and Kaiming He.
\newblock Exploring simple siamese representation learning.
\newblock In \emph{Proceedings of the IEEE/CVF conference on computer vision and pattern recognition}, pp.\  15750--15758, 2021.

\bibitem[Cheng et~al.(2021)Cheng, Hao, Yuan, Si, and Carin]{cheng2021fairfil}
Pengyu Cheng, Weituo Hao, Siyang Yuan, Shijing Si, and Lawrence Carin.
\newblock Fairfil: Contrastive neural debiasing method for pretrained text encoders.
\newblock \emph{arXiv preprint arXiv:2103.06413}, 2021.

\bibitem[Child et~al.()Child, Gray, Radford, and Sutskever]{child1904generating}
R~Child, S~Gray, A~Radford, and I~Sutskever.
\newblock Generating long sequences with sparse transformers. arxiv 2019.
\newblock \emph{arXiv preprint arXiv:1904.10509}.

\bibitem[Chuang et~al.(2020)Chuang, Robinson, Lin, Torralba, and Jegelka]{chuang2020debiased}
Ching-Yao Chuang, Joshua Robinson, Yen-Chen Lin, Antonio Torralba, and Stefanie Jegelka.
\newblock Debiased contrastive learning.
\newblock \emph{Advances in neural information processing systems}, 33:\penalty0 8765--8775, 2020.

\bibitem[Clark et~al.(2019)Clark, Khandelwal, Levy, and Manning]{clark-etal-2019-bert}
Kevin Clark, Urvashi Khandelwal, Omer Levy, and Christopher~D. Manning.
\newblock What does {BERT} look at? an analysis of {BERT}{'}s attention.
\newblock In \emph{Proceedings of the 2019 ACL Workshop BlackboxNLP: Analyzing and Interpreting Neural Networks for NLP}, pp.\  276--286, Florence, Italy, August 2019. Association for Computational Linguistics.
\newblock \doi{10.18653/v1/W19-4828}.
\newblock URL \url{https://www.aclweb.org/anthology/W19-4828}.

\bibitem[Creager et~al.(2019)Creager, Madras, Jacobsen, Weis, Swersky, Pitassi, and Zemel]{creager2019flexibly}
Elliot Creager, David Madras, J{\"o}rn-Henrik Jacobsen, Marissa Weis, Kevin Swersky, Toniann Pitassi, and Richard Zemel.
\newblock Flexibly fair representation learning by disentanglement.
\newblock In \emph{International conference on machine learning}, pp.\  1436--1445. PMLR, 2019.

\bibitem[Garg et~al.(2020)Garg, Villasenor, and Foggo]{garg2020fairness}
Pratyush Garg, John Villasenor, and Virginia Foggo.
\newblock Fairness metrics: A comparative analysis.
\newblock In \emph{2020 IEEE International Conference on Big Data (Big Data)}, pp.\  3662--3666. IEEE, 2020.

\bibitem[Hardt et~al.(2016)Hardt, Price, and Srebro]{hardt2016equality}
Moritz Hardt, Eric Price, and Nati Srebro.
\newblock Equality of opportunity in supervised learning.
\newblock \emph{Advances in neural information processing systems}, 29, 2016.

\bibitem[He et~al.(2016)He, Zhang, Ren, and Sun]{he2016deep}
Kaiming He, Xiangyu Zhang, Shaoqing Ren, and Jian Sun.
\newblock Deep residual learning for image recognition.
\newblock In \emph{Proceedings of the IEEE conference on computer vision and pattern recognition}, pp.\  770--778, 2016.

\bibitem[He et~al.(2020)He, Fan, Wu, Xie, and Girshick]{he2020momentum}
Kaiming He, Haoqi Fan, Yuxin Wu, Saining Xie, and Ross Girshick.
\newblock Momentum contrast for unsupervised visual representation learning.
\newblock In \emph{Proceedings of the IEEE/CVF conference on computer vision and pattern recognition}, pp.\  9729--9738, 2020.

\bibitem[Hewitt \& Liang(2019)Hewitt and Liang]{hewitt-liang-2019-designing}
John Hewitt and Percy Liang.
\newblock Designing and interpreting probes with control tasks.
\newblock In \emph{Proceedings of the 2019 Conference on Empirical Methods in Natural Language Processing and the 9th International Joint Conference on Natural Language Processing (EMNLP-IJCNLP)}, pp.\  2733--2743, Hong Kong, China, November 2019. Association for Computational Linguistics.
\newblock \doi{10.18653/v1/D19-1275}.
\newblock URL \url{https://www.aclweb.org/anthology/D19-1275}.

\bibitem[Hjelm et~al.(2018)Hjelm, Fedorov, Lavoie-Marchildon, Grewal, Bachman, Trischler, and Bengio]{hjelm2018learning}
R~Devon Hjelm, Alex Fedorov, Samuel Lavoie-Marchildon, Karan Grewal, Phil Bachman, Adam Trischler, and Yoshua Bengio.
\newblock Learning deep representations by mutual information estimation and maximization.
\newblock \emph{arXiv preprint arXiv:1808.06670}, 2018.

\bibitem[Hong \& Yang(2021)Hong and Yang]{hong2021unbiased}
Youngkyu Hong and Eunho Yang.
\newblock Unbiased classification through bias-contrastive and bias-balanced learning.
\newblock \emph{Advances in Neural Information Processing Systems}, 34:\penalty0 26449--26461, 2021.

\bibitem[Jung et~al.(2022)Jung, Chun, and Moon]{jung2022learning}
Sangwon Jung, Sanghyuk Chun, and Taesup Moon.
\newblock Learning fair classifiers with partially annotated group labels.
\newblock In \emph{Proceedings of the IEEE/CVF Conference on Computer Vision and Pattern Recognition}, pp.\  10348--10357, 2022.

\bibitem[Khajehnejad et~al.(2022)Khajehnejad, Khajehnejad, Babaei, Gummadi, Weller, and Mirzasoleiman]{khajehnejad2022crosswalk}
Ahmad Khajehnejad, Moein Khajehnejad, Mahmoudreza Babaei, Krishna~P Gummadi, Adrian Weller, and Baharan Mirzasoleiman.
\newblock Crosswalk: Fairness-enhanced node representation learning.
\newblock In \emph{Proceedings of the AAAI Conference on Artificial Intelligence}, volume~36, pp.\  11963--11970, 2022.

\bibitem[Kirk et~al.(2021)Kirk, Jun, Volpin, Iqbal, Benussi, Dreyer, Shtedritski, and Asano]{kirk2021bias}
Hannah~Rose Kirk, Yennie Jun, Filippo Volpin, Haider Iqbal, Elias Benussi, Frederic Dreyer, Aleksandar Shtedritski, and Yuki Asano.
\newblock Bias out-of-the-box: An empirical analysis of intersectional occupational biases in popular generative language models.
\newblock \emph{Advances in neural information processing systems}, 34:\penalty0 2611--2624, 2021.

\bibitem[Kitaev et~al.(2020)Kitaev, Kaiser, and Levskaya]{kitaev2020reformer}
Nikita Kitaev, {\L}ukasz Kaiser, and Anselm Levskaya.
\newblock Reformer: The efficient transformer.
\newblock \emph{arXiv preprint arXiv:2001.04451}, 2020.

\bibitem[Ling et~al.(2022)Ling, Jiang, Luo, Ji, and Zou]{ling2022learning}
Hongyi Ling, Zhimeng Jiang, Youzhi Luo, Shuiwang Ji, and Na~Zou.
\newblock Learning fair graph representations via automated data augmentations.
\newblock In \emph{The Eleventh International Conference on Learning Representations}, 2022.

\bibitem[Liu et~al.(2018)Liu, Luo, Wang, and Tang]{liu2018large}
Ziwei Liu, Ping Luo, Xiaogang Wang, and Xiaoou Tang.
\newblock Large-scale celebfaces attributes (celeba) dataset.
\newblock \emph{Retrieved August}, 15\penalty0 (2018):\penalty0 11, 2018.

\bibitem[Louizos et~al.(2015)Louizos, Swersky, Li, Welling, and Zemel]{louizos2015variational}
Christos Louizos, Kevin Swersky, Yujia Li, Max Welling, and Richard Zemel.
\newblock The variational fair autoencoder.
\newblock \emph{arXiv preprint arXiv:1511.00830}, 2015.

\bibitem[Lv et~al.(2023)Lv, Zhang, Zhang, Kuang, Wang, Wang, Chen, Shen, Yang, Ooi, et~al.]{lv2023duet}
Zheqi Lv, Wenqiao Zhang, Shengyu Zhang, Kun Kuang, Feng Wang, Yongwei Wang, Zhengyu Chen, Tao Shen, Hongxia Yang, Beng~Chin Ooi, et~al.
\newblock Duet: A tuning-free device-cloud collaborative parameters generation framework for efficient device model generalization.
\newblock In \emph{Proceedings of the ACM Web Conference 2023}, pp.\  3077--3085, 2023.

\bibitem[Madras et~al.(2018)Madras, Creager, Pitassi, and Zemel]{madras2018learning}
David Madras, Elliot Creager, Toniann Pitassi, and Richard Zemel.
\newblock Learning adversarially fair and transferable representations.
\newblock In \emph{International Conference on Machine Learning}, pp.\  3384--3393. PMLR, 2018.

\bibitem[Oord et~al.(2018)Oord, Li, and Vinyals]{oord2018representation}
Aaron van~den Oord, Yazhe Li, and Oriol Vinyals.
\newblock Representation learning with contrastive predictive coding.
\newblock \emph{arXiv preprint arXiv:1807.03748}, 2018.

\bibitem[Park et~al.(2022)Park, Lee, Lee, Hwang, Kim, and Byun]{park2022fair}
Sungho Park, Jewook Lee, Pilhyeon Lee, Sunhee Hwang, Dohyung Kim, and Hyeran Byun.
\newblock Fair contrastive learning for facial attribute classification.
\newblock In \emph{Proceedings of the IEEE/CVF Conference on Computer Vision and Pattern Recognition}, pp.\  10389--10398, 2022.

\bibitem[Parzen(1962)]{parzen1962estimation}
Emanuel Parzen.
\newblock On estimation of a probability density function and mode.
\newblock \emph{The annals of mathematical statistics}, 33\penalty0 (3):\penalty0 1065--1076, 1962.

\bibitem[Raff \& Sylvester(2018)Raff and Sylvester]{raff2018gradient}
Edward Raff and Jared Sylvester.
\newblock Gradient reversal against discrimination: A fair neural network learning approach.
\newblock In \emph{2018 IEEE 5th International Conference on Data Science and Advanced Analytics (DSAA)}, pp.\  189--198. IEEE, 2018.

\bibitem[Robinson et~al.(2020)Robinson, Chuang, Sra, and Jegelka]{robinson2020contrastive}
Joshua Robinson, Ching-Yao Chuang, Suvrit Sra, and Stefanie Jegelka.
\newblock Contrastive learning with hard negative samples.
\newblock \emph{arXiv preprint arXiv:2010.04592}, 2020.

\bibitem[Rosenblatt(1956)]{rosenblatt1956remarks}
Murray Rosenblatt.
\newblock Remarks on some nonparametric estimates of a density function.
\newblock \emph{The annals of mathematical statistics}, pp.\  832--837, 1956.

\bibitem[Sagawa et~al.(2019)Sagawa, Koh, Hashimoto, and Liang]{sagawa2019distributionally}
Shiori Sagawa, Pang~Wei Koh, Tatsunori~B Hashimoto, and Percy Liang.
\newblock Distributionally robust neural networks for group shifts: On the importance of regularization for worst-case generalization.
\newblock \emph{arXiv preprint arXiv:1911.08731}, 2019.

\bibitem[Shen et~al.(2021)Shen, Han, Cohn, Baldwin, and Frermann]{shen2021contrastive}
Aili Shen, Xudong Han, Trevor Cohn, Timothy Baldwin, and Lea Frermann.
\newblock Contrastive learning for fair representations.
\newblock \emph{arXiv preprint arXiv:2109.10645}, 2021.

\bibitem[Shrivastava \& Li(2014)Shrivastava and Li]{shrivastava2014asymmetric}
Anshumali Shrivastava and Ping Li.
\newblock Asymmetric lsh (alsh) for sublinear time maximum inner product search (mips).
\newblock \emph{Advances in neural information processing systems}, 27, 2014.

\bibitem[Song et~al.(2021)Song, Jung, Kim, and Moon]{song2021implicit}
Kyungwoo Song, Yohan Jung, Dongjun Kim, and Il-Chul Moon.
\newblock Implicit kernel attention.
\newblock In \emph{Proceedings of the AAAI Conference on Artificial Intelligence}, volume~35, pp.\  9713--9721, 2021.

\bibitem[Song et~al.(2013)Song, Fukumizu, and Gretton]{song2013kernel}
Le~Song, Kenji Fukumizu, and Arthur Gretton.
\newblock Kernel embeddings of conditional distributions: A unified kernel framework for nonparametric inference in graphical models.
\newblock \emph{IEEE Signal Processing Magazine}, 30\penalty0 (4):\penalty0 98--111, 2013.

\bibitem[Tao et~al.(2023)Tao, Tonin, Patrinos, and Suykens]{tao2023nonlinear}
Qinghua Tao, Francesco Tonin, Panagiotis Patrinos, and Johan~AK Suykens.
\newblock Nonlinear svd with asymmetric kernels: feature learning and asymmetric nystr$\backslash$" om method.
\newblock \emph{arXiv preprint arXiv:2306.07040}, 2023.

\bibitem[Tenney et~al.(2019)Tenney, Das, and Pavlick]{tenney-etal-2019-bert}
Ian Tenney, Dipanjan Das, and Ellie Pavlick.
\newblock {BERT} rediscovers the classical {NLP} pipeline.
\newblock In \emph{Proceedings of the 57th Annual Meeting of the Association for Computational Linguistics}, pp.\  4593--4601, Florence, Italy, July 2019. Association for Computational Linguistics.
\newblock \doi{10.18653/v1/P19-1452}.
\newblock URL \url{https://www.aclweb.org/anthology/P19-1452}.

\bibitem[Tian et~al.(2020)Tian, Sun, Poole, Krishnan, Schmid, and Isola]{tian2020makes}
Yonglong Tian, Chen Sun, Ben Poole, Dilip Krishnan, Cordelia Schmid, and Phillip Isola.
\newblock What makes for good views for contrastive learning?
\newblock \emph{Advances in neural information processing systems}, 33:\penalty0 6827--6839, 2020.

\bibitem[Tsai et~al.(2019)Tsai, Bai, Yamada, Morency, and Salakhutdinov]{tsai2019transformer}
Yao-Hung~Hubert Tsai, Shaojie Bai, Makoto Yamada, Louis-Philippe Morency, and Ruslan Salakhutdinov.
\newblock Transformer dissection: a unified understanding of transformer's attention via the lens of kernel.
\newblock \emph{arXiv preprint arXiv:1908.11775}, 2019.

\bibitem[Tsai et~al.(2021{\natexlab{a}})Tsai, Ma, Yang, Zhao, Morency, and Salakhutdinov]{tsai2021self}
Yao-Hung~Hubert Tsai, Martin~Q Ma, Muqiao Yang, Han Zhao, Louis-Philippe Morency, and Ruslan Salakhutdinov.
\newblock Self-supervised representation learning with relative predictive coding.
\newblock \emph{arXiv preprint arXiv:2103.11275}, 2021{\natexlab{a}}.

\bibitem[Tsai et~al.(2021{\natexlab{b}})Tsai, Ma, Zhao, Zhang, Morency, and Salakhutdinov]{tsai2021conditional}
Yao-Hung~Hubert Tsai, Martin~Q Ma, Han Zhao, Kun Zhang, Louis-Philippe Morency, and Ruslan Salakhutdinov.
\newblock Conditional contrastive learning: Removing undesirable information in self-supervised representations.
\newblock \emph{arXiv e-prints}, pp.\  arXiv--2106, 2021{\natexlab{b}}.

\bibitem[Tsai et~al.(2022)Tsai, Li, Ma, Zhao, Zhang, Morency, and Salakhutdinov]{tsai2022conditional}
Yao-Hung~Hubert Tsai, Tianqin Li, Martin~Q Ma, Han Zhao, Kun Zhang, Louis-Philippe Morency, and Ruslan Salakhutdinov.
\newblock Conditional contrastive learning with kernel.
\newblock \emph{arXiv preprint arXiv:2202.05458}, 2022.

\bibitem[Vaswani et~al.(2017)Vaswani, Shazeer, Parmar, Uszkoreit, Jones, Gomez, Kaiser, and Polosukhin]{vaswani2017attention}
Ashish Vaswani, Noam Shazeer, Niki Parmar, Jakob Uszkoreit, Llion Jones, Aidan~N Gomez, {\L}ukasz Kaiser, and Illia Polosukhin.
\newblock Attention is all you need.
\newblock \emph{Advances in neural information processing systems}, 30, 2017.

\bibitem[Vig \& Belinkov(2019)Vig and Belinkov]{vig-belinkov-2019-analyzing}
Jesse Vig and Yonatan Belinkov.
\newblock Analyzing the structure of attention in a transformer language model.
\newblock In \emph{Proceedings of the 2019 ACL Workshop BlackboxNLP: Analyzing and Interpreting Neural Networks for NLP}, pp.\  63--76, Florence, Italy, August 2019. Association for Computational Linguistics.
\newblock \doi{10.18653/v1/W19-4808}.
\newblock URL \url{https://www.aclweb.org/anthology/W19-4808}.

\bibitem[Voita et~al.(2019)Voita, Talbot, Moiseev, Sennrich, and Titov]{voita-etal-2019-analyzing}
Elena Voita, David Talbot, Fedor Moiseev, Rico Sennrich, and Ivan Titov.
\newblock Analyzing multi-head self-attention: Specialized heads do the heavy lifting, the rest can be pruned.
\newblock In \emph{Proceedings of the 57th Annual Meeting of the Association for Computational Linguistics}, pp.\  5797--5808, Florence, Italy, July 2019. Association for Computational Linguistics.
\newblock \doi{10.18653/v1/P19-1580}.
\newblock URL \url{https://www.aclweb.org/anthology/P19-1580}.

\bibitem[Wang et~al.(2019)Wang, Zhao, Yatskar, Chang, and Ordonez]{wang2019balanced}
Tianlu Wang, Jieyu Zhao, Mark Yatskar, Kai-Wei Chang, and Vicente Ordonez.
\newblock Balanced datasets are not enough: Estimating and mitigating gender bias in deep image representations.
\newblock In \emph{Proceedings of the IEEE/CVF international conference on computer vision}, pp.\  5310--5319, 2019.

\bibitem[Wright \& Gonzalez(2021)Wright and Gonzalez]{wright2021transformers}
Matthew~A Wright and Joseph~E Gonzalez.
\newblock Transformers are deep infinite-dimensional non-mercer binary kernel machines.
\newblock \emph{arXiv preprint arXiv:2106.01506}, 2021.

\bibitem[Wu et~al.(2020)Wu, Mosse, Zhuang, Yamins, and Goodman]{wu2020conditional}
Mike Wu, Milan Mosse, Chengxu Zhuang, Daniel Yamins, and Noah Goodman.
\newblock Conditional negative sampling for contrastive learning of visual representations.
\newblock \emph{arXiv preprint arXiv:2010.02037}, 2020.

\bibitem[You et~al.(2017)You, Gitman, and Ginsburg]{you2017large}
Yang You, Igor Gitman, and Boris Ginsburg.
\newblock Large batch training of convolutional networks.
\newblock \emph{arXiv preprint arXiv:1708.03888}, 2017.

\bibitem[Zhang et~al.(2023)Zhang, Cen, and Shah]{zhang2023matrix}
Cindy Zhang, Sarah~Huiyi Cen, and Devavrat Shah.
\newblock Matrix estimation for individual fairness.
\newblock In \emph{International Conference on Machine Learning}, pp.\  40871--40887. PMLR, 2023.

\bibitem[Zhang et~al.(2022)Zhang, Kuang, Chen, Liu, Wu, and Xiao]{zhang2022fairness}
Fengda Zhang, Kun Kuang, Long Chen, Yuxuan Liu, Chao Wu, and Jun Xiao.
\newblock Fairness-aware contrastive learning with partially annotated sensitive attributes.
\newblock In \emph{The Eleventh International Conference on Learning Representations}, 2022.

\end{thebibliography}
